%% file: main.tex
\documentclass[]{article}
\usepackage{graphicx}
\usepackage{tabularx}
\usepackage[dvipsnames,table]{xcolor}
\usepackage[font=scriptsize,labelfont=bf]{caption}
\usepackage{amsfonts}
\usepackage{amsmath}
\usepackage{cite}
\usepackage{multirow}
\usepackage{multicol}
\usepackage{booktabs}
\usepackage{graphicx} % Required for inserting images
\usepackage{titlesec}
\usepackage{hyperref}
\usepackage[square,numbers]{natbib}
\titlespacing{\section}{0pc}{1pc}{1pc}

\title{Detecting Dataset Bias in Medical AI: A Generalized and Modality-Agnostic Auditing Framework}
\author{
  Nathan Drenkow$^{\ast1, 2}$ \and
  Mitchell Pavlak$^{\ast1}$ \and
  Keith Harrigian$^1$ \and
  Ayah Zirikly$^1$ \and \newline
  Adarsh Subbaswamy$^3$ \and
  Mohammad Mehdi Farhangi$^3$ \and
  Nicholas Petrick$^3$ \and
  Mathias Unberath$^1$  \\ 
  {\small $^1$ The Johns Hopkins University} \\
  {\small $^2$ The Johns Hopkins University Applied Physics Laboratory} \\
  {\small $^3$ Center for Devices and Radiological Health, U.S. Food and Drug Administration} \\
  {\small $^\ast$ Equal contribution}
}
\date{}

\begin{document}
\input{math_commands}

\maketitle

\begin{abstract}
Artificial Intelligence (AI) is now firmly at the center of evidence-based medicine. Despite many success stories that edge the path of AI's rise in healthcare, there are comparably many reports of significant shortcomings and unexpected behavior of AI in deployment. 
A major reason for these limitations is AI's reliance on association-based learning, where non-representative machine learning datasets can amplify latent bias during training and/or hide it during testing. To unlock new tools capable of foreseeing and preventing such AI bias issues, we present G-AUDIT. Generalized Attribute Utility and Detectability-Induced bias Testing (G-AUDIT) for datasets is a modality-agnostic auditing framework that allows for generating targeted hypotheses about sources of bias in training or testing data.  
Our method examines the relationship between task-level annotations (commonly referred to as ``labels'') and data properties including patient attributes (e.g., age, sex) and environment and acquisition characteristics (e.g., clinical site, imaging protocols). 
G-AUDIT automatically quantifies the extent to which the observed data attributes pose a risk for shortcut learning, or in the case of testing data, might hide predictions made based on spurious associations. 
We demonstrate the broad applicability and value of our method by analyzing large-scale medical datasets for three distinct modalities and machine learning tasks: skin lesion classification in images, stigmatizing language classification in Electronic Health Records (EHR), and mortality prediction for ICU tabular data. 
In each setting, G-AUDIT successfully identifies subtle biases commonly overlooked by traditional qualitative methods, underscoring its practical value in exposing dataset-level risks and supporting the downstream development of reliable AI systems.
% Our method contributes to achieving a deeper understanding of machine learning datasets throughout the AI development life-cycle from initial prototyping all the way to deployment 
% and creates opportunities for reducing AI model bias, enabling safer and more trustworthy AI systems. 
\end{abstract}

\input{sections/Intro}
\input{sections/Results}

\input{sections/Discussion}

\input{sections/Methods}

\section{Data Availability}
Data for the ICU mortality prediction (\url{https://mimic.mit.edu}) and skin lesion classification cases (\url{https://challenge.isic-archive.com/data}) are publicly available. Because clinical notes from the Johns Hopkins Medicine (JHM) dataset contain identifiable information, they cannot be shared outside of our study team; however, all code to replicate the experiment on other datasets will be made available.  Data was maintained, controlled and accessed only by Johns Hopkins University and not by the U.S. Food and Drug Administration.

\section{Acknowledgment}
MP acknowledges partial funding by appointment to the Research Participation Program at the Center for Devices and Radiological Health administered by the Oak Ridge Institute for Science and Education through an interagency agreement between the US Department of Energy and the US Food and Drug Administration. The mention of commercial products, their sources, or their use in connection with material reported herein is not to be construed as either an actual or implied endorsement of such products by the Department of Health and Human Services. This is a contribution of the U.S. Food and Drug Administration and is not subject to copyright 

\bibliography{egbib}
\bibliographystyle{abbrvnat}

\clearpage

\input{sections/Supplemental}

\end{document}

%% file: math_commands.tex
%%%%% NEW MATH DEFINITIONS %%%%%

% Mark sections of captions for referring to divisions of figures
\newcommand{\figleft}{{\em (Left)}}
\newcommand{\figcenter}{{\em (Center)}}
\newcommand{\figright}{{\em (Right)}}
\newcommand{\figtop}{{\em (Top)}}
\newcommand{\figbottom}{{\em (Bottom)}}
\newcommand{\captiona}{{\em (a)}}
\newcommand{\captionb}{{\em (b)}}
\newcommand{\captionc}{{\em (c)}}
\newcommand{\captiond}{{\em (d)}}

% Highlight a newly defined term
\newcommand{\newterm}[1]{{\bf #1}}

% Figure reference, lower-case.
\def\figref#1{figure~\ref{#1}}
% Figure reference, capital. For start of sentence
\def\Figref#1{Figure~\ref{#1}}
\def\twofigref#1#2{figures \ref{#1} and \ref{#2}}
\def\quadfigref#1#2#3#4{figures \ref{#1}, \ref{#2}, \ref{#3} and \ref{#4}}
% Section reference, lower-case.
\def\secref#1{section~\ref{#1}}
% Section reference, capital.
\def\Secref#1{Section~\ref{#1}}
% Reference to two sections.
\def\twosecrefs#1#2{sections \ref{#1} and \ref{#2}}
% Reference to three sections.
\def\secrefs#1#2#3{sections \ref{#1}, \ref{#2} and \ref{#3}}
% Reference to an equation, lower-case.
\def\eqref#1{equation~\ref{#1}}
% Reference to an equation, upper case
\def\Eqref#1{Equation~\ref{#1}}
% A raw reference to an equation---avoid using if possible
\def\plaineqref#1{\ref{#1}}
% Reference to a chapter, lower-case.
\def\chapref#1{chapter~\ref{#1}}
% Reference to an equation, upper case.
\def\Chapref#1{Chapter~\ref{#1}}
% Reference to a range of chapters
\def\rangechapref#1#2{chapters\ref{#1}--\ref{#2}}
% Reference to an algorithm, lower-case.
\def\algref#1{algorithm~\ref{#1}}
% Reference to an algorithm, upper case.
\def\Algref#1{Algorithm~\ref{#1}}
\def\twoalgref#1#2{algorithms \ref{#1} and \ref{#2}}
\def\Twoalgref#1#2{Algorithms \ref{#1} and \ref{#2}}
% Reference to a part, lower case
\def\partref#1{part~\ref{#1}}
% Reference to a part, upper case
\def\Partref#1{Part~\ref{#1}}
\def\twopartref#1#2{parts \ref{#1} and \ref{#2}}

\def\ceil#1{\lceil #1 \rceil}
\def\floor#1{\lfloor #1 \rfloor}
\def\1{\bm{1}}
\newcommand{\train}{\mathcal{D}}
\newcommand{\valid}{\mathcal{D_{\mathrm{valid}}}}
\newcommand{\test}{\mathcal{D_{\mathrm{test}}}}

\def\eps{{\epsilon}}

% Random variables
\def\reta{{\textnormal{$\eta$}}}
\def\ra{{\textnormal{a}}}
\def\rb{{\textnormal{b}}}
\def\rc{{\textnormal{c}}}
\def\rd{{\textnormal{d}}}
\def\re{{\textnormal{e}}}
\def\rf{{\textnormal{f}}}
\def\rg{{\textnormal{g}}}
\def\rh{{\textnormal{h}}}
\def\ri{{\textnormal{i}}}
\def\rj{{\textnormal{j}}}
\def\rk{{\textnormal{k}}}
\def\rl{{\textnormal{l}}}
% rm is already a command, just don't name any random variables m
\def\rn{{\textnormal{n}}}
\def\ro{{\textnormal{o}}}
\def\rp{{\textnormal{p}}}
\def\rq{{\textnormal{q}}}
\def\rr{{\textnormal{r}}}
\def\rs{{\textnormal{s}}}
\def\rt{{\textnormal{t}}}
\def\ru{{\textnormal{u}}}
\def\rv{{\textnormal{v}}}
\def\rw{{\textnormal{w}}}
\def\rx{{\textnormal{x}}}
\def\ry{{\textnormal{y}}}
\def\rz{{\textnormal{z}}}

% Random vectors
\def\rvepsilon{{\mathbf{\epsilon}}}
\def\rvtheta{{\mathbf{\theta}}}
\def\rva{{\mathbf{a}}}
\def\rvb{{\mathbf{b}}}
\def\rvc{{\mathbf{c}}}
\def\rvd{{\mathbf{d}}}
\def\rve{{\mathbf{e}}}
\def\rvf{{\mathbf{f}}}
\def\rvg{{\mathbf{g}}}
\def\rvh{{\mathbf{h}}}
\def\rvu{{\mathbf{i}}}
\def\rvj{{\mathbf{j}}}
\def\rvk{{\mathbf{k}}}
\def\rvl{{\mathbf{l}}}
\def\rvm{{\mathbf{m}}}
\def\rvn{{\mathbf{n}}}
\def\rvo{{\mathbf{o}}}
\def\rvp{{\mathbf{p}}}
\def\rvq{{\mathbf{q}}}
\def\rvr{{\mathbf{r}}}
\def\rvs{{\mathbf{s}}}
\def\rvt{{\mathbf{t}}}
\def\rvu{{\mathbf{u}}}
\def\rvv{{\mathbf{v}}}
\def\rvw{{\mathbf{w}}}
\def\rvx{{\mathbf{x}}}
\def\rvy{{\mathbf{y}}}
\def\rvz{{\mathbf{z}}}

% Elements of random vectors
\def\erva{{\textnormal{a}}}
\def\ervb{{\textnormal{b}}}
\def\ervc{{\textnormal{c}}}
\def\ervd{{\textnormal{d}}}
\def\erve{{\textnormal{e}}}
\def\ervf{{\textnormal{f}}}
\def\ervg{{\textnormal{g}}}
\def\ervh{{\textnormal{h}}}
\def\ervi{{\textnormal{i}}}
\def\ervj{{\textnormal{j}}}
\def\ervk{{\textnormal{k}}}
\def\ervl{{\textnormal{l}}}
\def\ervm{{\textnormal{m}}}
\def\ervn{{\textnormal{n}}}
\def\ervo{{\textnormal{o}}}
\def\ervp{{\textnormal{p}}}
\def\ervq{{\textnormal{q}}}
\def\ervr{{\textnormal{r}}}
\def\ervs{{\textnormal{s}}}
\def\ervt{{\textnormal{t}}}
\def\ervu{{\textnormal{u}}}
\def\ervv{{\textnormal{v}}}
\def\ervw{{\textnormal{w}}}
\def\ervx{{\textnormal{x}}}
\def\ervy{{\textnormal{y}}}
\def\ervz{{\textnormal{z}}}

% Random matrices
\def\rmA{{\mathbf{A}}}
\def\rmB{{\mathbf{B}}}
\def\rmC{{\mathbf{C}}}
\def\rmD{{\mathbf{D}}}
\def\rmE{{\mathbf{E}}}
\def\rmF{{\mathbf{F}}}
\def\rmG{{\mathbf{G}}}
\def\rmH{{\mathbf{H}}}
\def\rmI{{\mathbf{I}}}
\def\rmJ{{\mathbf{J}}}
\def\rmK{{\mathbf{K}}}
\def\rmL{{\mathbf{L}}}
\def\rmM{{\mathbf{M}}}
\def\rmN{{\mathbf{N}}}
\def\rmO{{\mathbf{O}}}
\def\rmP{{\mathbf{P}}}
\def\rmQ{{\mathbf{Q}}}
\def\rmR{{\mathbf{R}}}
\def\rmS{{\mathbf{S}}}
\def\rmT{{\mathbf{T}}}
\def\rmU{{\mathbf{U}}}
\def\rmV{{\mathbf{V}}}
\def\rmW{{\mathbf{W}}}
\def\rmX{{\mathbf{X}}}
\def\rmY{{\mathbf{Y}}}
\def\rmZ{{\mathbf{Z}}}

% Elements of random matrices
\def\ermA{{\textnormal{A}}}
\def\ermB{{\textnormal{B}}}
\def\ermC{{\textnormal{C}}}
\def\ermD{{\textnormal{D}}}
\def\ermE{{\textnormal{E}}}
\def\ermF{{\textnormal{F}}}
\def\ermG{{\textnormal{G}}}
\def\ermH{{\textnormal{H}}}
\def\ermI{{\textnormal{I}}}
\def\ermJ{{\textnormal{J}}}
\def\ermK{{\textnormal{K}}}
\def\ermL{{\textnormal{L}}}
\def\ermM{{\textnormal{M}}}
\def\ermN{{\textnormal{N}}}
\def\ermO{{\textnormal{O}}}
\def\ermP{{\textnormal{P}}}
\def\ermQ{{\textnormal{Q}}}
\def\ermR{{\textnormal{R}}}
\def\ermS{{\textnormal{S}}}
\def\ermT{{\textnormal{T}}}
\def\ermU{{\textnormal{U}}}
\def\ermV{{\textnormal{V}}}
\def\ermW{{\textnormal{W}}}
\def\ermX{{\textnormal{X}}}
\def\ermY{{\textnormal{Y}}}
\def\ermZ{{\textnormal{Z}}}

% Vectors
\def\vzero{{\bm{0}}}
\def\vone{{\bm{1}}}
\def\vmu{{\bm{\mu}}}
\def\vtheta{{\bm{\theta}}}
\def\va{{\bm{a}}}
\def\vb{{\bm{b}}}
\def\vc{{\bm{c}}}
\def\vd{{\bm{d}}}
\def\ve{{\bm{e}}}
\def\vf{{\bm{f}}}
\def\vg{{\bm{g}}}
\def\vh{{\bm{h}}}
\def\vi{{\bm{i}}}
\def\vj{{\bm{j}}}
\def\vk{{\bm{k}}}
\def\vl{{\bm{l}}}
\def\vm{{\bm{m}}}
\def\vn{{\bm{n}}}
\def\vo{{\bm{o}}}
\def\vp{{\bm{p}}}
\def\vq{{\bm{q}}}
\def\vr{{\bm{r}}}
\def\vs{{\bm{s}}}
\def\vt{{\bm{t}}}
\def\vu{{\bm{u}}}
\def\vv{{\bm{v}}}
\def\vw{{\bm{w}}}
\def\vx{{\bm{x}}}
\def\vy{{\bm{y}}}
\def\vz{{\bm{z}}}

% Elements of vectors
\def\evalpha{{\alpha}}
\def\evbeta{{\beta}}
\def\evepsilon{{\epsilon}}
\def\evlambda{{\lambda}}
\def\evomega{{\omega}}
\def\evmu{{\mu}}
\def\evpsi{{\psi}}
\def\evsigma{{\sigma}}
\def\evtheta{{\theta}}
\def\eva{{a}}
\def\evb{{b}}
\def\evc{{c}}
\def\evd{{d}}
\def\eve{{e}}
\def\evf{{f}}
\def\evg{{g}}
\def\evh{{h}}
\def\evi{{i}}
\def\evj{{j}}
\def\evk{{k}}
\def\evl{{l}}
\def\evm{{m}}
\def\evn{{n}}
\def\evo{{o}}
\def\evp{{p}}
\def\evq{{q}}
\def\evr{{r}}
\def\evs{{s}}
\def\evt{{t}}
\def\evu{{u}}
\def\evv{{v}}
\def\evw{{w}}
\def\evx{{x}}
\def\evy{{y}}
\def\evz{{z}}

% Matrix
\def\mA{{\bm{A}}}
\def\mB{{\bm{B}}}
\def\mC{{\bm{C}}}
\def\mD{{\bm{D}}}
\def\mE{{\bm{E}}}
\def\mF{{\bm{F}}}
\def\mG{{\bm{G}}}
\def\mH{{\bm{H}}}
\def\mI{{\bm{I}}}
\def\mJ{{\bm{J}}}
\def\mK{{\bm{K}}}
\def\mL{{\bm{L}}}
\def\mM{{\bm{M}}}
\def\mN{{\bm{N}}}
\def\mO{{\bm{O}}}
\def\mP{{\bm{P}}}
\def\mQ{{\bm{Q}}}
\def\mR{{\bm{R}}}
\def\mS{{\bm{S}}}
\def\mT{{\bm{T}}}
\def\mU{{\bm{U}}}
\def\mV{{\bm{V}}}
\def\mW{{\bm{W}}}
\def\mX{{\bm{X}}}
\def\mY{{\bm{Y}}}
\def\mZ{{\bm{Z}}}
\def\mBeta{{\bm{\beta}}}
\def\mPhi{{\bm{\Phi}}}
\def\mLambda{{\bm{\Lambda}}}
\def\mSigma{{\bm{\Sigma}}}

% Tensor
% \DeclareMathAlphabet{\mathsfit}{\encodingdefault}{\sfdefault}{m}{sl}
% \SetMathAlphabet{\mathsfit}{bold}{\encodingdefault}{\sfdefault}{bx}{n}
\newcommand{\tens}[1]{\bm{\mathsfit{#1}}}
\def\tA{{\tens{A}}}
\def\tB{{\tens{B}}}
\def\tC{{\tens{C}}}
\def\tD{{\tens{D}}}
\def\tE{{\tens{E}}}
\def\tF{{\tens{F}}}
\def\tG{{\tens{G}}}
\def\tH{{\tens{H}}}
\def\tI{{\tens{I}}}
\def\tJ{{\tens{J}}}
\def\tK{{\tens{K}}}
\def\tL{{\tens{L}}}
\def\tM{{\tens{M}}}
\def\tN{{\tens{N}}}
\def\tO{{\tens{O}}}
\def\tP{{\tens{P}}}
\def\tQ{{\tens{Q}}}
\def\tR{{\tens{R}}}
\def\tS{{\tens{S}}}
\def\tT{{\tens{T}}}
\def\tU{{\tens{U}}}
\def\tV{{\tens{V}}}
\def\tW{{\tens{W}}}
\def\tX{{\tens{X}}}
\def\tY{{\tens{Y}}}
\def\tZ{{\tens{Z}}}

% Graph
\def\gA{{\mathcal{A}}}
\def\gB{{\mathcal{B}}}
\def\gC{{\mathcal{C}}}
\def\gD{{\mathcal{D}}}
\def\gE{{\mathcal{E}}}
\def\gF{{\mathcal{F}}}
\def\gG{{\mathcal{G}}}
\def\gH{{\mathcal{H}}}
\def\gI{{\mathcal{I}}}
\def\gJ{{\mathcal{J}}}
\def\gK{{\mathcal{K}}}
\def\gL{{\mathcal{L}}}
\def\gM{{\mathcal{M}}}
\def\gN{{\mathcal{N}}}
\def\gO{{\mathcal{O}}}
\def\gP{{\mathcal{P}}}
\def\gQ{{\mathcal{Q}}}
\def\gR{{\mathcal{R}}}
\def\gS{{\mathcal{S}}}
\def\gT{{\mathcal{T}}}
\def\gU{{\mathcal{U}}}
\def\gV{{\mathcal{V}}}
\def\gW{{\mathcal{W}}}
\def\gX{{\mathcal{X}}}
\def\gY{{\mathcal{Y}}}
\def\gZ{{\mathcal{Z}}}

% Sets
\def\sA{{\mathbb{A}}}
\def\sB{{\mathbb{B}}}
\def\sC{{\mathbb{C}}}
\def\sD{{\mathbb{D}}}
% Don't use a set called E, because this would be the same as our symbol
% for expectation.
\def\sF{{\mathbb{F}}}
\def\sG{{\mathbb{G}}}
\def\sH{{\mathbb{H}}}
\def\sI{{\mathbb{I}}}
\def\sJ{{\mathbb{J}}}
\def\sK{{\mathbb{K}}}
\def\sL{{\mathbb{L}}}
\def\sM{{\mathbb{M}}}
\def\sN{{\mathbb{N}}}
\def\sO{{\mathbb{O}}}
\def\sP{{\mathbb{P}}}
\def\sQ{{\mathbb{Q}}}
\def\sR{{\mathbb{R}}}
\def\sS{{\mathbb{S}}}
\def\sT{{\mathbb{T}}}
\def\sU{{\mathbb{U}}}
\def\sV{{\mathbb{V}}}
\def\sW{{\mathbb{W}}}
\def\sX{{\mathbb{X}}}
\def\sY{{\mathbb{Y}}}
\def\sZ{{\mathbb{Z}}}

% Entries of a matrix
\def\emLambda{{\Lambda}}
\def\emA{{A}}
\def\emB{{B}}
\def\emC{{C}}
\def\emD{{D}}
\def\emE{{E}}
\def\emF{{F}}
\def\emG{{G}}
\def\emH{{H}}
\def\emI{{I}}
\def\emJ{{J}}
\def\emK{{K}}
\def\emL{{L}}
\def\emM{{M}}
\def\emN{{N}}
\def\emO{{O}}
\def\emP{{P}}
\def\emQ{{Q}}
\def\emR{{R}}
\def\emS{{S}}
\def\emT{{T}}
\def\emU{{U}}
\def\emV{{V}}
\def\emW{{W}}
\def\emX{{X}}
\def\emY{{Y}}
\def\emZ{{Z}}
\def\emSigma{{\Sigma}}

% entries of a tensor
% Same font as tensor, without \bm wrapper
\newcommand{\etens}[1]{\mathsfit{#1}}
\def\etLambda{{\etens{\Lambda}}}
\def\etA{{\etens{A}}}
\def\etB{{\etens{B}}}
\def\etC{{\etens{C}}}
\def\etD{{\etens{D}}}
\def\etE{{\etens{E}}}
\def\etF{{\etens{F}}}
\def\etG{{\etens{G}}}
\def\etH{{\etens{H}}}
\def\etI{{\etens{I}}}
\def\etJ{{\etens{J}}}
\def\etK{{\etens{K}}}
\def\etL{{\etens{L}}}
\def\etM{{\etens{M}}}
\def\etN{{\etens{N}}}
\def\etO{{\etens{O}}}
\def\etP{{\etens{P}}}
\def\etQ{{\etens{Q}}}
\def\etR{{\etens{R}}}
\def\etS{{\etens{S}}}
\def\etT{{\etens{T}}}
\def\etU{{\etens{U}}}
\def\etV{{\etens{V}}}
\def\etW{{\etens{W}}}
\def\etX{{\etens{X}}}
\def\etY{{\etens{Y}}}
\def\etZ{{\etens{Z}}}

% The true underlying data generating distribution
\newcommand{\pdata}{p_{\rm{data}}}
% The empirical distribution defined by the training set
\newcommand{\ptrain}{\hat{p}_{\rm{data}}}
\newcommand{\Ptrain}{\hat{P}_{\rm{data}}}
% The model distribution
\newcommand{\pmodel}{p_{\rm{model}}}
\newcommand{\Pmodel}{P_{\rm{model}}}
\newcommand{\ptildemodel}{\tilde{p}_{\rm{model}}}
% Stochastic autoencoder distributions
\newcommand{\pencode}{p_{\rm{encoder}}}
\newcommand{\pdecode}{p_{\rm{decoder}}}
\newcommand{\precons}{p_{\rm{reconstruct}}}

\newcommand{\laplace}{\mathrm{Laplace}} % Laplace distribution

\newcommand{\E}{\mathbb{E}}
\newcommand{\Ls}{\mathcal{L}}
\newcommand{\R}{\mathbb{R}}
\newcommand{\emp}{\tilde{p}}
\newcommand{\lr}{\alpha}
\newcommand{\reg}{\lambda}
\newcommand{\rect}{\mathrm{rectifier}}
\newcommand{\softmax}{\mathrm{softmax}}
\newcommand{\sigmoid}{\sigma}
\newcommand{\softplus}{\zeta}
\newcommand{\KL}{D_{\mathrm{KL}}}
\newcommand{\Var}{\mathrm{Var}}
\newcommand{\standarderror}{\mathrm{SE}}
\newcommand{\Cov}{\mathrm{Cov}}
% Wolfram Mathworld says $L^2$ is for function spaces and $\ell^2$ is for vectors
% But then they seem to use $L^2$ for vectors throughout the site, and so does
% wikipedia.
\newcommand{\normlzero}{L^0}
\newcommand{\normlone}{L^1}
\newcommand{\normltwo}{L^2}
\newcommand{\normlp}{L^p}
\newcommand{\normmax}{L^\infty}

\newcommand{\parents}{Pa} % See usage in notation.tex. Chosen to match Daphne's book.

\newcommand{\argmax}{arg\,max}
\newcommand{\argmin}{arg\,min}

\newcommand{\sign}{sign}
\newcommand{\Tr}{Tr}
\let\ab\allowbreak

%% file: sections/Intro.tex
\section{Introduction}
The use of artificial intelligence (AI) and machine learning (ML) in healthcare 
has created numerous opportunities to improve quality, safety, efficiency, efficacy, affordability, and accessibility of healthcare through applications including disease detection~\cite{liu2019comparison}, forecasting~\cite{bertsimas2021machine,herbert2023forecasting}, opportunistic screening~\cite{pickhardt2021opportunistic}, and quantitative medicine~\cite{dreizin2022pilot}. However, AI/ML models have repeatedly exhibited deterioration in performance and implicit biases when moving from the development phase to testing and deployment phases~\cite{wong2021external,matsoukas2023ai, drenkow2021systematic, Jaspers2024-vv, Ong_Ly2024-jh, geirhos2020shortcut, o2022evaluating}. Due to the data-driven nature of AI/ML methods, the brittleness and bias of these models is inherently tied to the data on which they are trained and evaluated prior to deployment.  Bias in medical datasets may originate from a number of sources~\cite{Jones2024-na, Gichoya2023-gf} and the growing size of these datasets~\cite{Chambon2024-vz, Johnson2023-mf, johnson2016mimic} increases the risk that latent biases may go undetected during data collection and then subsequently exploited by downstream AI/ML models. 

The influence of dataset biases may manifest in a number of ways including shortcut learning~\cite{winkler2019association, winkler2021association, nauta2021uncovering, jabbour2020deep, degrave2021ai, Ong_Ly2024-jh}, bias and fairness issues in AI predictions~\cite{Gichoya2022ReadingRace, Seyyed-Kalantari2021-aj, Seyyed-Kalantari2021-fa, Glocker2022-st, Henry_Hinnefeld2018-vi}, and lack of generalization or robustness~\cite{o2022evaluating}. All of these issues are problematic, as they may result in harmful and unpredictable performance disparities of the model within and across patient populations in deployment.   
The best opportunity to address data-driven AI/ML bias risks starts with exposing \emph{dataset-level risks} prior to model training or evaluation.  Therefore, dataset audits, the focus of this work, are of critical importance for identifying and quantifying model bias as early as possible in the training and evaluation phases.

\begin{figure}[b!]
    \centering
\includegraphics[width=\textwidth]{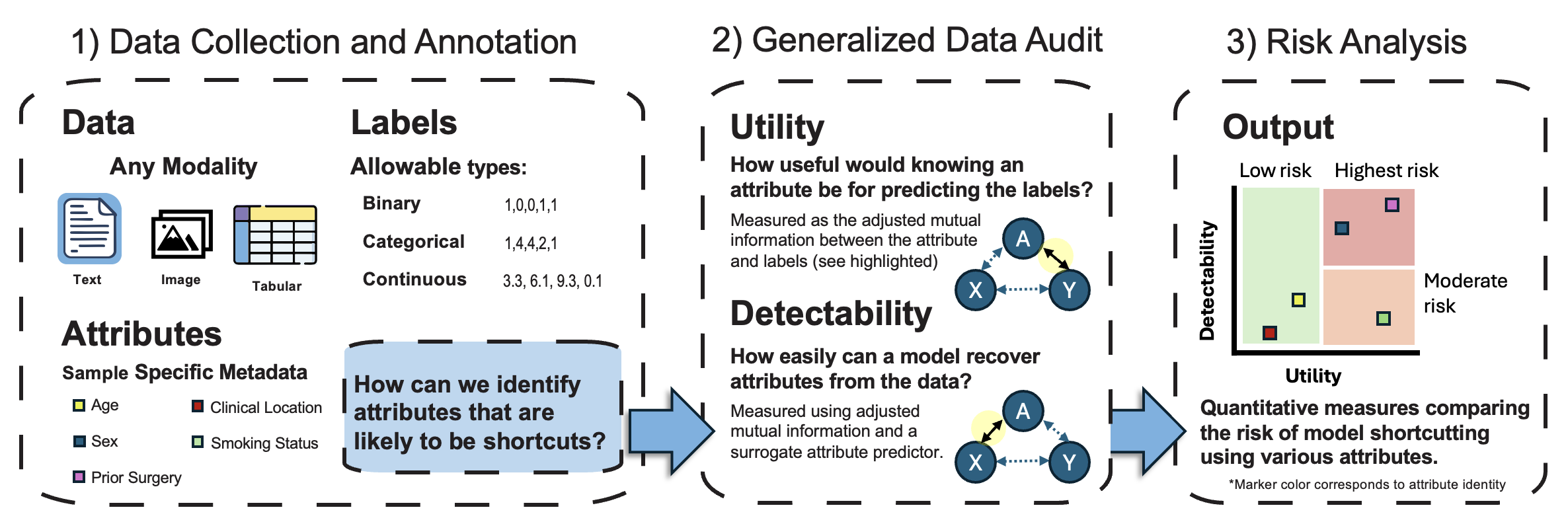}
    \caption{Given a dataset consisting of input with associated labels and metadata, we perform our Generalized Attribute Utility and Detectability-Induced bias Testing (G-AUDIT) to generate quantitative hypotheses on the relative risk of attributes in the form of a Detectability vs Utility scatter plot (right).}
\label{fig:gda_overview}
\end{figure}

Despite the importance of quantifying bias risk at the dataset-level, techniques for performing dataset audits remain largely absent~\cite{Jones2024-na}. On the one hand, some methods may examine disparities in medical dataset metadata~\cite{Henry_Hinnefeld2018-vi, Aka2021-kq, oakden2020hidden} but they do not explicitly link these to the primary data with which AI/ML models are trained.  In contrast, other methods show that patient attributes may be predictable from sensor-level measurements~\cite{Gichoya2022ReadingRace, glocker2023algorithmic} but may not fully identify whether the attribute-level features represent direct shortcuts for achieving accurate prediction. Some methods more directly examine the relation between data attributes and task model accuracy~\cite{Seyyed-Kalantari2021-aj, Seyyed-Kalantari2021-fa, jabbour2020deep, Subbaswamy2024-cr} yet these do not enable assessment of the dataset-level risks independent of task-level modeling assumptions. Lastly, while most methods have focused on addressing bias with respect to conventional patient attributes (such as age, sex), some methods have begun to explore biases that may relate to other aspects of the data collection pipeline such as image acquisition or clinician-level markings~\cite{Ong_Ly2024-jh, Pavlak2023-km}.  In short, current dataset auditing methods lack practical utility because they do not holistically address \emph{both} the risk of relevant attributes leaking information about the task \emph{and} the risk of whether information about those attributes can be directly exploited during model training.  Furthermore, existing auditing methods have been tailored to specific modalities or clinical tasks leading to a paucity of modality-agnostic methods for quantifying dataset-level bias risks. Thus, new automated dataset auditing techniques are required to identify these primary risks and generate hypotheses to guide downstream model development, auditing, and mitigation.

To address the need for quantitative dataset audits, we present a novel technique that represents a significant advance in enabling independent, dataset-level auditing and which we refer to as Generalized Attribute Utility and Detectability-Induced bias Testing (G-AUDIT) for datasets.  Our method presents the first unified approach to shortcut auditing by considering the interplay between attribute-level composition, sensor-level measurements, and task labels.  G-AUDIT is generalizable and not tied to a particular data modality or ML task.
Our procedure (Figure~\ref{fig:gda_overview}) utilizes the tools of information theory and causal inference to identify the presence and strength of association between attributes and task labels (\emph{utility}) as well as the ability to infer the values of those attributes directly from the data itself (\emph{detectability}). We then rank attributes according to their measured utility and detectability, thus providing AI/ML developers and clinicians alike a quantitative yet interpretable means of assessing dataset-level bias risk.
Our method is not limited to only patient-level attributes and allows for datasets to be audited with respect to any variables of relevance to the data collection process, patient population, and machine learning task.

We apply our G-AUDIT method to a diverse set of clinical tasks covering image, text, and tabular data modalities.  In each dataset, our audits uncover shortcut risks (confirmed by domain experts) linked to non-patient attributes representing aspects of the data collection pipeline not typically considered as candidates for bias. We also provide an optional calibration procedure which allows for estimating a form of worst case performance degradation given the observed values of utility for attributes in the dataset. For the potential shortcuts flagged by our auditing method, we find that the estimated worst case degradation is substantial (often $>0.2$ Area Under the Curve difference) indicating that models exploiting these shortcuts may stand to exhibit significant drops in performance in deployment. By applying G-AUDIT across clinical tasks and data modalities, our results underscore and address the need for new methods to quantitatively assess dataset bias risk across all aspects of the data collection process. 

%% file: sections/Results.tex
\section{Results}
To demonstrate its broad effectiveness, our G-AUDIT procedure is applied to datasets from three different modalities, namely, image, text, and tabular data.  For each dataset, we estimate the utility and detectability (see Sec.~\ref{sec:methods}) of each attribute relative to the underlying learning task. We do this to generate hypotheses that identify and rank potential shortcut attributes according to the risk that downstream models may exploit them. We also include results from an optional calibration procedure in which an approximate upper bound on the drop in performance-related metrics can be calculated for each potential shortcut using a synthetic attribute in precisely controlled conditions (see Section~\ref{subsec:bounding_performance}). The bound provides a means of estimating a form of worst-case downstream model performance risks for specific attributes in more familiar metrics.

\subsection{Skin Lesion Classification}
We first consider shortcut risks in vision-based tasks. The high dimensionality of image data creates many opportunities for shortcuts to exist without directly impacting the machine learning task.  For instance, hospital-specific tokens placed in the chest X-ray field of view (e.g.,~\cite{zech2018variable}) may impact only a small number of pixels in the image but may constitute a shortcut when associated primarily with a specific disease condition in the training dataset. In cases like these, deep neural networks (DNNs) trained on the data may exploit salient features like tokens/watermarks (or other statistical regularities not task-related) to achieve low training error. We focus here on skin lesion classification where the construction of large-scale datasets may not be able to adequately balance across patient characteristics, clinical sites, and dermascopic imaging sensors/settings and where such features of the dataset and collection process may manifest as shortcuts.

The ISIC 2019 skin lesion dataset~\cite{tschandl2018ham10000, codella2018skin, combalia2019bcn20000} was analyzed for bias with respect to the included metadata and patient attributes. The dataset consists of 25,331 training images and attributes \verb|age|, \verb|race|, \verb|sex|, \verb|anatomical location|, and \verb|skin color| on the Fitzpatrick scale.  Image metadata includes \verb|height|, \verb|width|, and \verb|year| of collection. While the original task labels included seven diagnostic categories, we reduce the classification task to malignant vs. benign conditions (e.g.,~\cite{zhang2019attention, lopez2017skin, mahbod2019skin}).  The data auditing procedure (see Sec.~\ref{sec:methods}) was applied to the entire training dataset relative to the binary malignancy classification task.  Images with any missing attribute values were excluded from the analysis.

\begin{figure}
    \centering
    \includegraphics[width=0.48\textwidth]{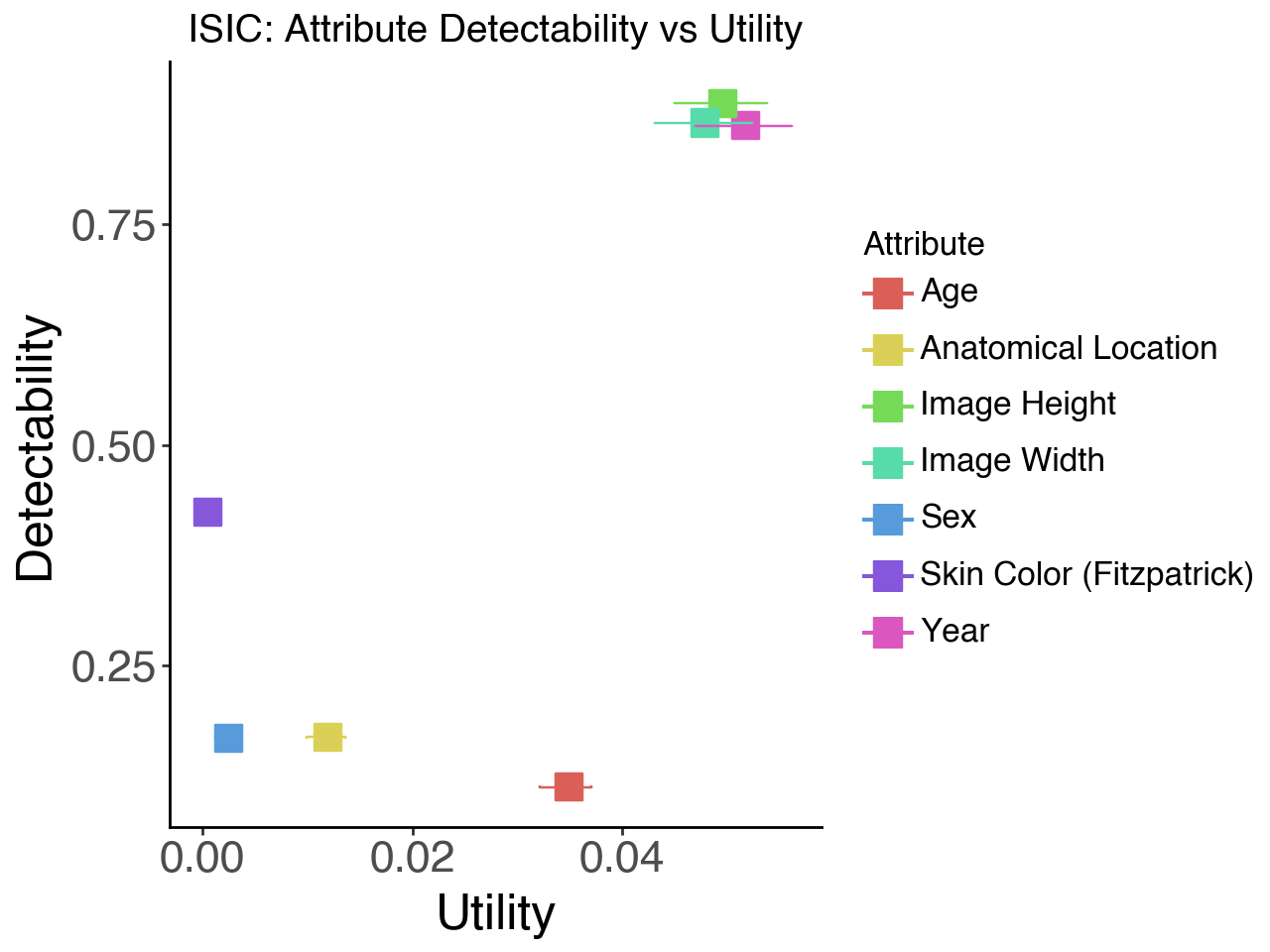}
    \includegraphics[width=0.48\linewidth]{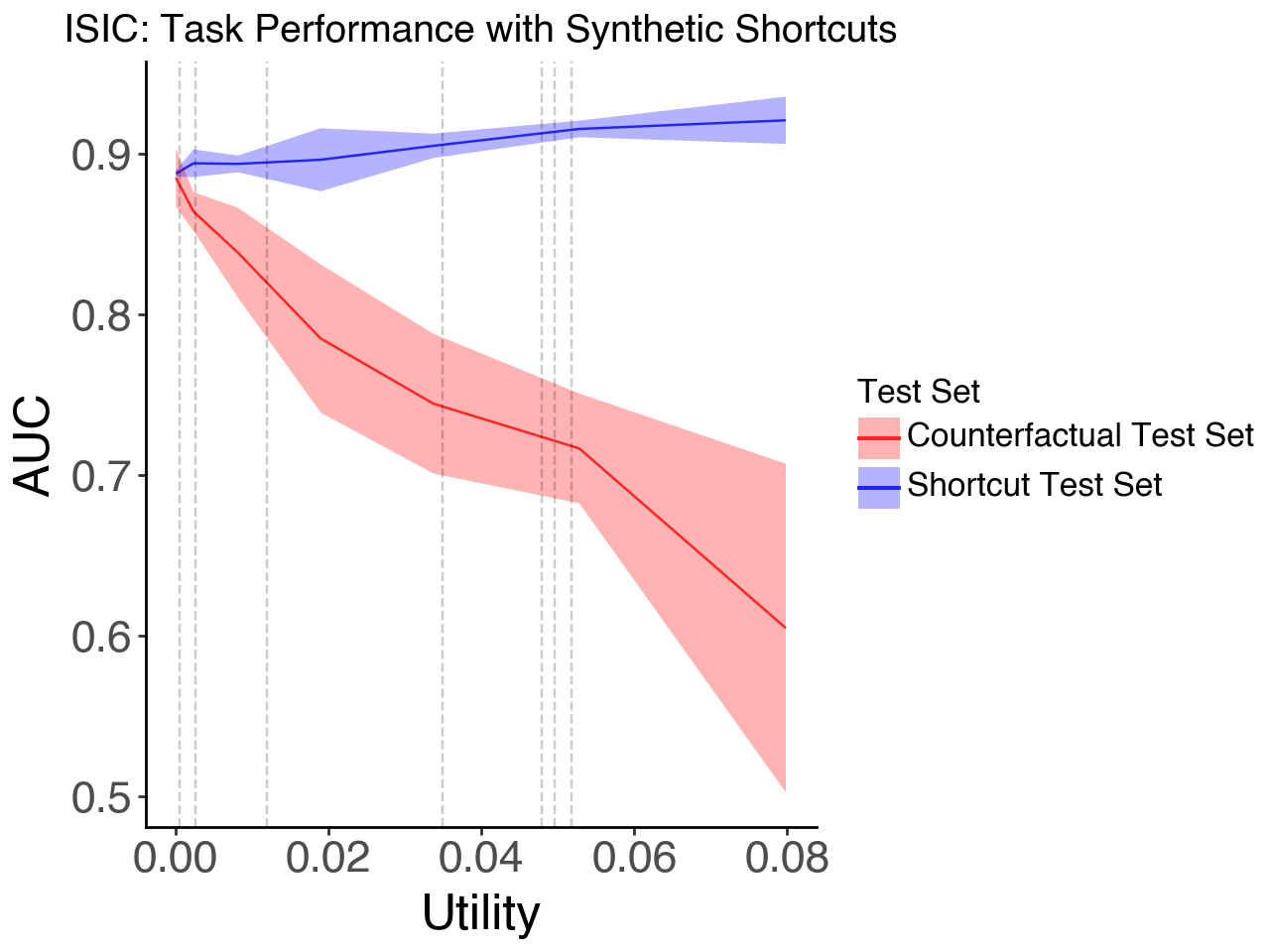}
    \caption{(left) Attribute Detectability vs. Utility for the ISIC 2019 dataset. Confidence intervals calculated via basic bootstrap \cite{efron1993bootstrap} (right) 
    Approximate worst-case performance risk for a synthetic attribute with varying utility (see Sec~\ref{subsec:bounding_performance} for more detail). Confidence intervals are derived from a t-distribution based on model performance across three folds; vertical lines indicate observed values of utility and indicate a worse-case drop in AUC of $\approx 0.2$.}
    \label{fig:isic-mi-cmi}
\end{figure}

The main auditing results are found in Figure~\ref{fig:isic-mi-cmi} and Table~\ref{tab:skin-utility-detectability} where the \verb|height|, \verb|width|, and \verb|year| attributes exhibit the highest combination of utility and detectability, indicating these attributes are more likely sources of bias within the dataset.  While these particular attributes may seem unrelated to the task labels, this image metadata can act as a proxy for the camera type and/or clinical site.  Furthermore, while all images are resized and cropped to a fixed resolution prior to running the auditing procedure, the high detectability scores indicate that some of this information is still retained in the images themselves, either directly or via proxy.

\begin{table}
\centering
\caption{Comparison between Utility and Detectability measures as well as macro-averaged F1 scores from the detection model and conducting the SPLIT procedure \cite{Gichoya2022ReadingRace, glocker2023algorithmic} on trained task models (see Sec.~\ref{subsub:split}).}
\resizebox{\linewidth}{!}{%
\begin{tabular}{lrr|rr}
\toprule
\bfseries  & \multicolumn{2}{c}{\bfseries{G-AUDIT (ours)}} & \multicolumn{2}{c}{\bfseries{Baselines}} \\
\bfseries Attribute & \bfseries Utility & \bfseries Detectability & \bfseries Detection F1 & \bfseries SPLIT F1 \\
\midrule
{\cellcolor[HTML]{F5F5F5}} \color[HTML]{000000} Year & {\cellcolor[HTML]{F5F5F5}} \color[HTML]{000000} 0.052 & {\cellcolor[HTML]{F5F5F5}} \color[HTML]{000000} 0.862 & {\cellcolor[HTML]{F5F5F5}} \color[HTML]{000000} 0.952 & {\cellcolor[HTML]{F5F5F5}} \color[HTML]{000000} 0.981 \\
{\cellcolor[HTML]{D3D3D3}} \color[HTML]{000000} Image Height & {\cellcolor[HTML]{D3D3D3}} \color[HTML]{000000} 0.050 & {\cellcolor[HTML]{D3D3D3}} \color[HTML]{000000} 0.887 & {\cellcolor[HTML]{D3D3D3}} \color[HTML]{000000} 0.918 & {\cellcolor[HTML]{D3D3D3}} \color[HTML]{000000} 0.948 \\
{\cellcolor[HTML]{F5F5F5}} \color[HTML]{000000} Image Width & {\cellcolor[HTML]{F5F5F5}} \color[HTML]{000000} 0.048 & {\cellcolor[HTML]{F5F5F5}} \color[HTML]{000000} 0.865 & {\cellcolor[HTML]{F5F5F5}} \color[HTML]{000000} 0.510 & {\cellcolor[HTML]{F5F5F5}} \color[HTML]{000000} 0.583 \\
{\cellcolor[HTML]{D3D3D3}} \color[HTML]{000000} Age & {\cellcolor[HTML]{D3D3D3}} \color[HTML]{000000} 0.035 & {\cellcolor[HTML]{D3D3D3}} \color[HTML]{000000} 0.112 & {\cellcolor[HTML]{D3D3D3}} \color[HTML]{000000} 0.292 & {\cellcolor[HTML]{D3D3D3}} \color[HTML]{000000} 0.334 \\
{\cellcolor[HTML]{F5F5F5}} \color[HTML]{000000} Anatomical Location & {\cellcolor[HTML]{F5F5F5}} \color[HTML]{000000} 0.012 & {\cellcolor[HTML]{F5F5F5}} \color[HTML]{000000} 0.169 & {\cellcolor[HTML]{F5F5F5}} \color[HTML]{000000} 0.288 & {\cellcolor[HTML]{F5F5F5}} \color[HTML]{000000} 0.624 \\
{\cellcolor[HTML]{D3D3D3}} \color[HTML]{000000} Sex & {\cellcolor[HTML]{D3D3D3}} \color[HTML]{000000} 0.003 & {\cellcolor[HTML]{D3D3D3}} \color[HTML]{000000} 0.168 & {\cellcolor[HTML]{D3D3D3}} \color[HTML]{000000} 0.736 & {\cellcolor[HTML]{D3D3D3}} \color[HTML]{000000} 0.768 \\
{\cellcolor[HTML]{F5F5F5}} \color[HTML]{000000} Skin Color (Fitzpatrick) & {\cellcolor[HTML]{F5F5F5}} \color[HTML]{000000} 0.000 & {\cellcolor[HTML]{F5F5F5}} \color[HTML]{000000} 0.424 & {\cellcolor[HTML]{F5F5F5}} \color[HTML]{000000} 0.538 & {\cellcolor[HTML]{F5F5F5}} \color[HTML]{000000} 0.632 \\
\bottomrule
\end{tabular}
}%
\label{tab:skin-utility-detectability}
\end{table}

\subsection{Stigmatizing Language in Electronic Health Records}
We next consider the text domain where the dimensionality of the data remains high and the variability of natural language create unique opportunities for shortcuts to exist. Similarly to the image domain, DNNs are often used to learn compact representations of natural language text for various tasks and may exploit shortcuts in the data that are not relevant to the clinical task yet allow them to achieve low training error. We apply the G-AUDIT procedure to the electronic medical record dataset introduced by \cite{harrigian2023characterization} for the purpose of characterizing stigmatizing language usage by physicians. The dataset contains 5,201 annotated instances across 3 tasks, with each task focusing on a different thematic group of stigmatizing language --- Credibility \& Obstinacy, Compliance, and Descriptions of Appearance/Demeanor. Models are provided a window of text centered around a keyword or phrase which has been identified by domain experts as being a potential indicator of unconscious bias towards a patient. They are asked to characterize the implication of the input instance (e.g., ``the patient \emph{claims} to brush their teeth 2x daily''; ``unable to track down insurance \emph{claims}''). Each instance is associated with auxiliary attributes which indicate a patient's race and gender, and the clinical setting from which the statement was drawn (e.g., OB-Gyn, Surgery).

The stigmatizing language dataset serves as an interesting case study for several reasons. First, the authors of \cite{harrigian2023characterization} included an analysis which examined how well each attribute could be predicted based on the last embedding layer of the primary stigmatizing language task models. Their results provide a direct reference for our method. Second, as is often true in practice, the stigmatizing language dataset has an ambiguous causal structure. This allows us to evaluate the robustness of our method to potential errors in misspecification of the direction of dependency between attributes and labels. Finally, there are documented disparities in the prevalence of stigmatizing language between demographic groups which we expect to show up directly within our utility measure (e.g., Black patients are more likely than white patients to experience discrimination). To gather an unbiased estimate of the prevalence of stigmatizing language in the population, models should not make use of demographics as a predictive shortcut.

\begin{figure}[h!]
    \centering
    \includegraphics[width=0.3\linewidth]{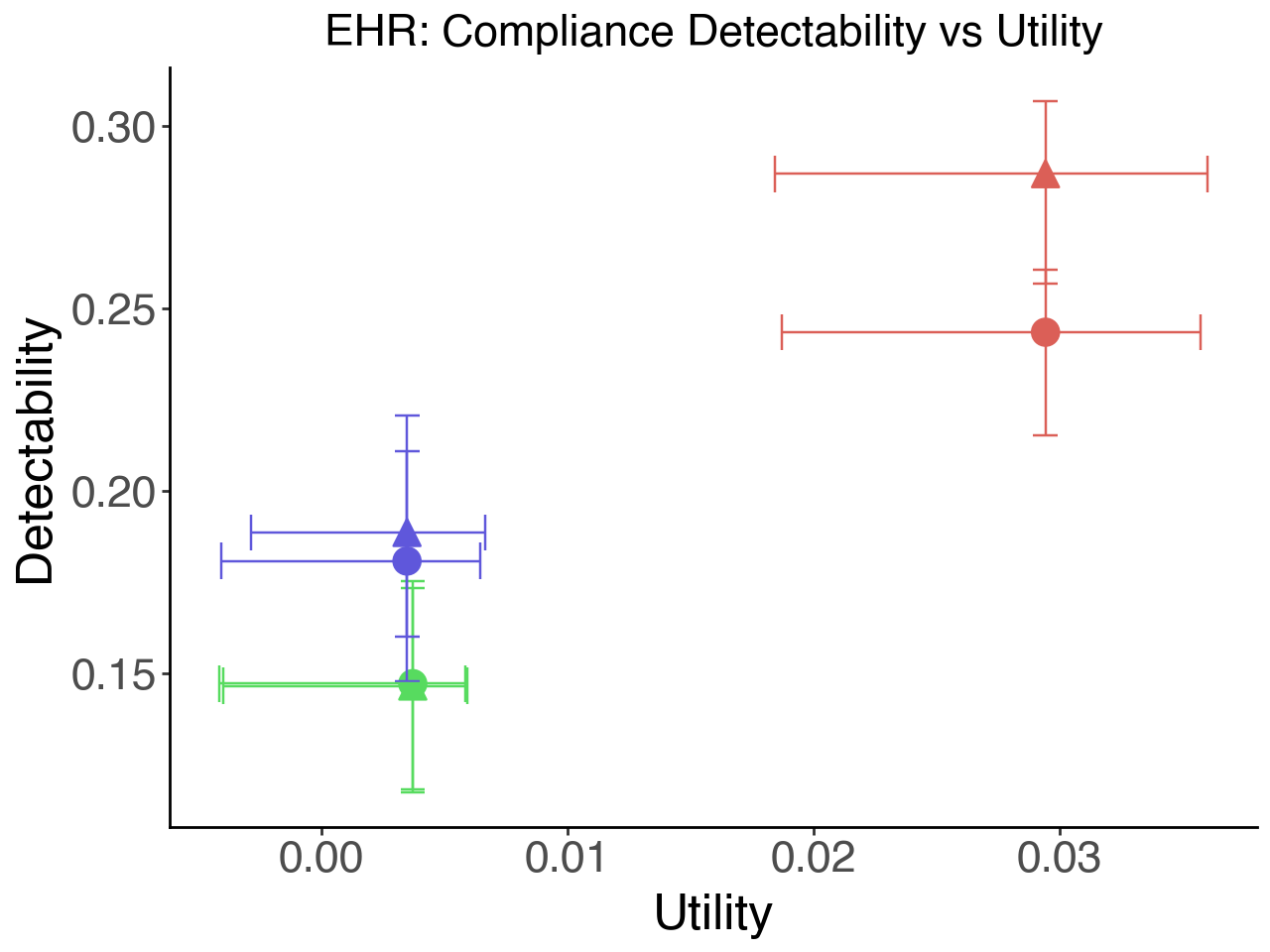}
    \includegraphics[width=0.3\linewidth]{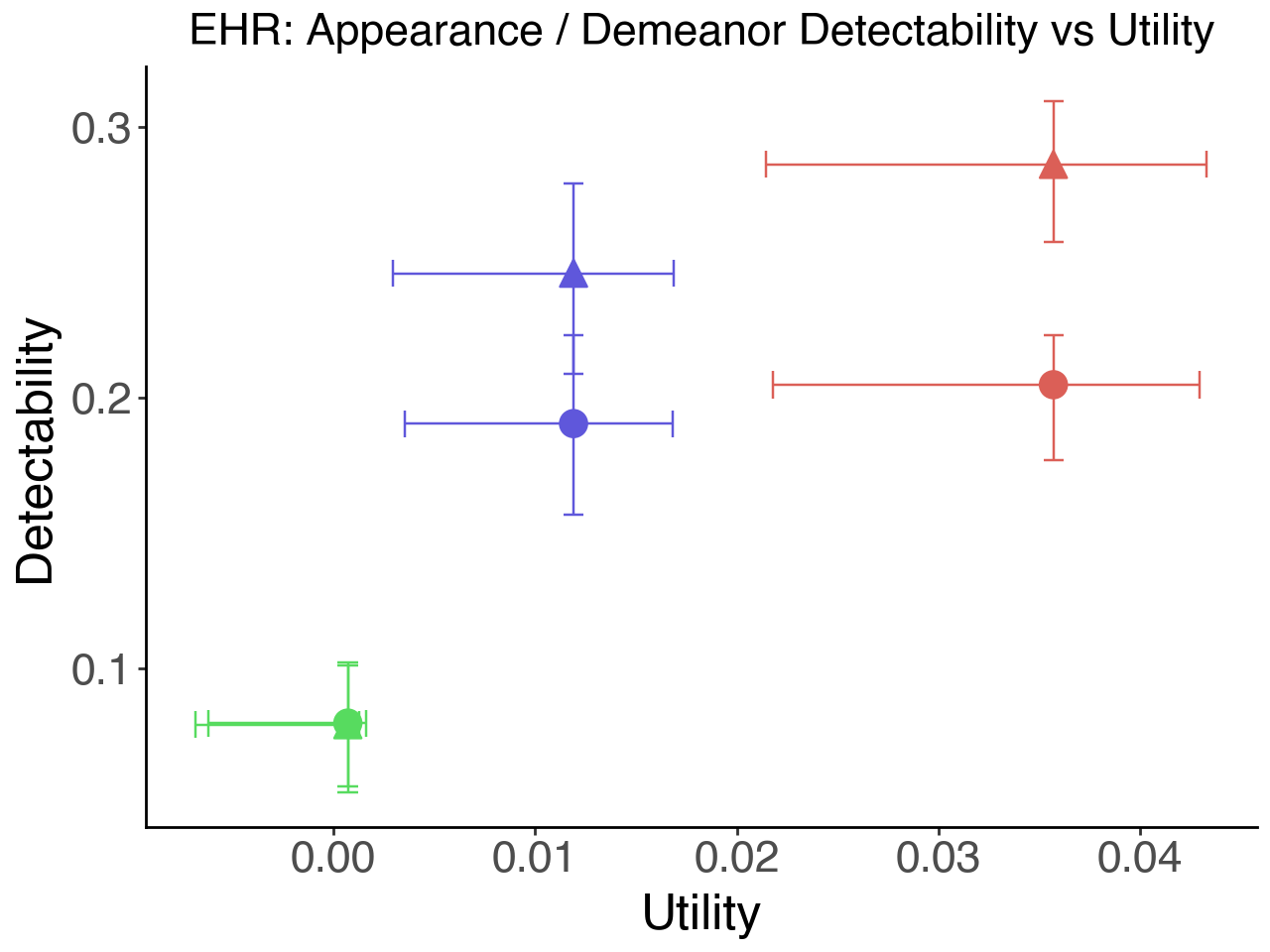}
    \includegraphics[width=0.3\linewidth]{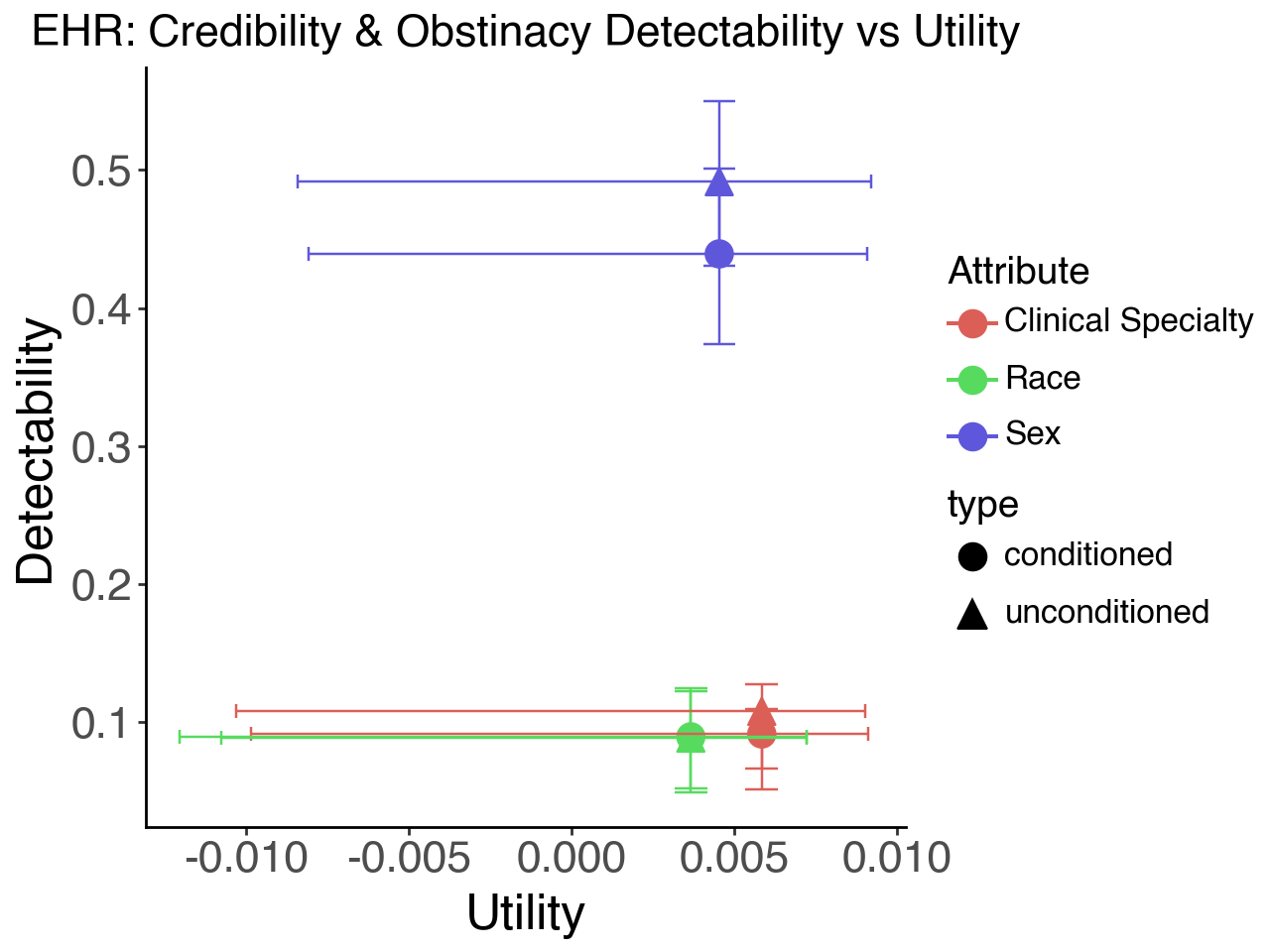}
    \caption{
     Utility vs. detectability for three identical attributes across three different EHR-related tasks when calculating detectability for the (unconditioned) $X \rightarrow Y$ and (conditioned) $Y \rightarrow X$ cases. 95\% confidence intervals calculated via basic bootstrap\cite{efron1993bootstrap}.}
    \label{fig:utility_detectability_cond}
\end{figure}

As shown in Figure~\ref{fig:utility_detectability_cond}, clinical specialty had a higher utility than both patient race and sex for the compliance and appearance / demeanor tasks.
Prior work shows that downstream models for these tasks likely did not encode sex and race characteristics beyond what could be explained by a reliance on clinical specialty alone \cite{harrigian2023characterization}. This is consistent with our observation that clinical specialty had a higher utility than race and sex, implying a stronger association with the task label.  
Importantly, this does not preclude performance disparities based on these sensitive attributes—clinical specialties in the JHM dataset include OB-GYN with an all-female patient population and Pediatrics with an approximately 95\% Black patient population \cite{harrigian2023characterization}. Instead, our results suggest that downstream models are more likely to exploit shortcuts related to the  identification of different clinical domains than to directly encode race or sex to improve performance. 

In terms of detectability, we find differences between conditioned and unconditioned measures to be fairly small. Interestingly, within the Credibility and Obstinacy task, while all attributes had relatively low utility, the detectability of sex was higher than that of any other EHR task and attribute evaluated. This presents a potential explanation for the findings of  \cite{harrigian2023characterization}, which identified sex within the Credibility and Obstinacy task as the only demographic attribute recoverable above baselines levels across all three tasks. 

\subsection{Mortality Prediction with Intensive Care Unit Tabular Data}
Lastly, we consider the dataset auditing problem in the context of tabular data.  While not as feature rich or high dimensional as images and text, tabular data presents its own unique challenges from an auditing perspective. 
While the features learned by task models in the image and text domains often lack interpretability (e.g., DNN embeddings), tabular data provides a direct mapping between attributes and their measured values.  In these cases,  attributes used as inputs to a task model can still act as shortcuts when they exhibit undesirable associations with the task labels that could arise due to sample bias or incorrect usage of the attribute itself.
For instance, if clinical sites differ in their test-ordering protocols, associations between disease conditions and clinical site may be reflected in the test orders/results provided to machine learning task models~\cite{Subbaswamy2021-su}. Furthermore, the gap in task performance between DNNs and more traditional ML models (e.g., SVMs, ensemble methods, etc.) is much smaller in the tabular case and this provides an opportunity to measure both detectability and task model performance over a wider class of machine learning models.

Here, we evaluated our auditing methodology on tabular data extracted from the publicly available Medical Information Mart for Intensive Care (MIMIC) III Dataset \cite{johnson2016mimic}. Specifically, we extracted a dataset for predicting mortality in Intensive Care Unit (ICU) patients using features involved in the Simplified Acute Physiology II (SAPS II) score \cite{gallPhysiology1993,pirracchio2015mortality}, a score used to measure disease severity in patients after their first 24 hours in the ICU. These features include patient age as well as summary statistics of heart rate, systolic blood pressure, temperature, labwork indicators/results, ventilator usage, and Glasgow Coma Scale (GCS) during the first 24 hours of the patient's ICU stay.

Our task is to predict patient mortality based on the tabularized data of a given patient. The final processed dataset consists of 34,386 patient records and 40 features, detailed in Appendix~\ref{subsec:shortcuts}. Features are one-hot encoded and include medication details, missing variables (e.g., GCS or lab test results), and patient demographics such as race and insurance coverage. We select an FT-Transformer~\cite{gorishniy2021FtTransformer} with default hyperparameters for use as a task model in all experiments. A benefit of this selection is the ability to apply the SPLIT approach \cite{glocker2023algorithmic,Gichoya2022ReadingRace} to empirically test how well our detectability measure corresponds with other baseline approaches for measuring detectability \cite{glocker2023algorithmic,Gichoya2022ReadingRace}. 

\begin{figure}[h]
    \centering
\includegraphics[width=.8\textwidth]{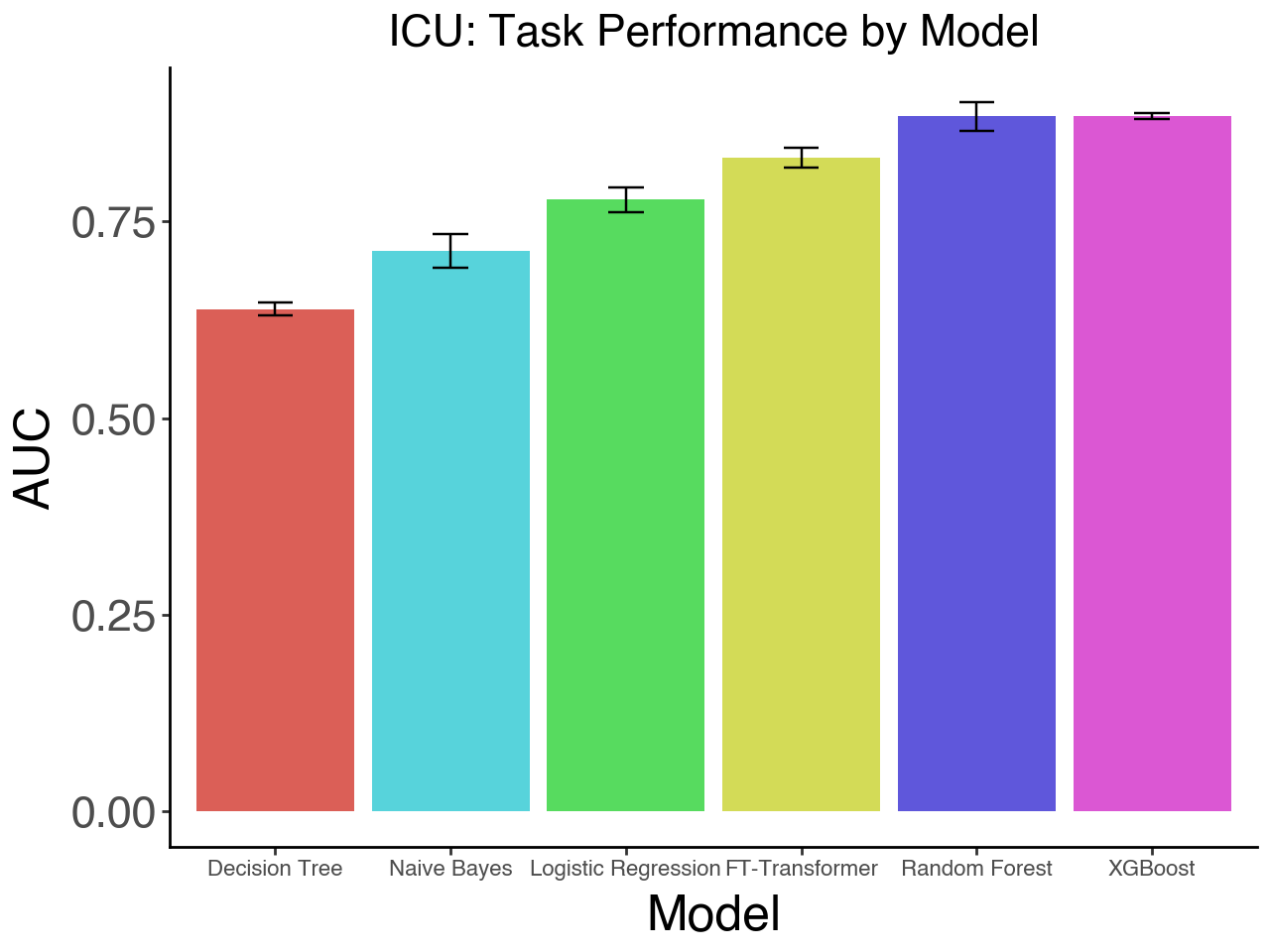}
    \caption{We first calibrate expectations by exploring the unmodified performance of various models on the task of predicting mortality.}
\label{fig:sepsis_task_performance}
\end{figure}

\begin{figure}[h]
    \centering
\includegraphics[width=.8\textwidth]{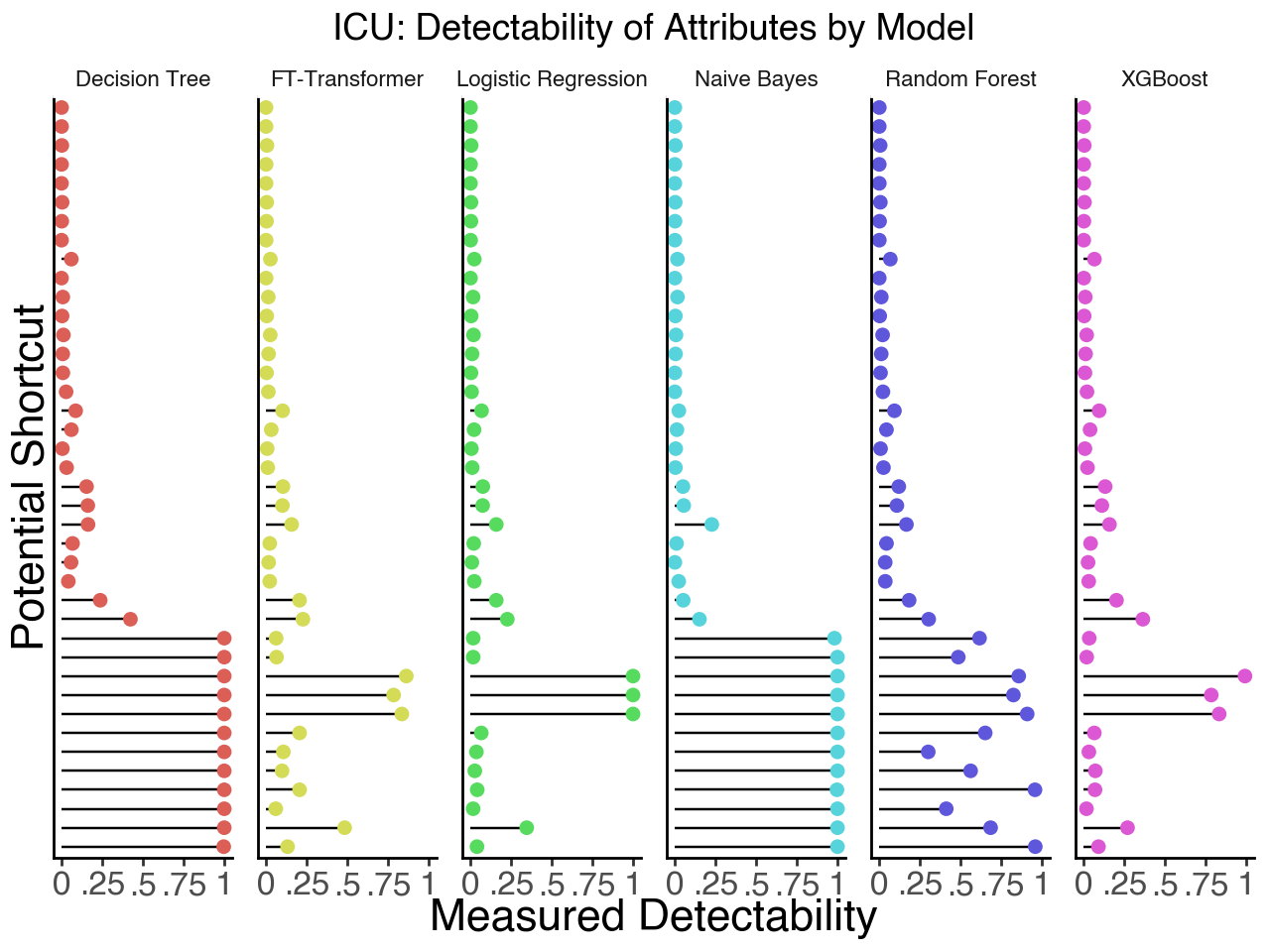}
    \caption{We order each attribute from least to greatest recoverability in a downstream FT-Transformer task model as measured by SPLIT~\cite{glocker2023algorithmic,Gichoya2022ReadingRace}.  By doing this, we see that attributes that are more recoverable tend to have higher detectability values, though there is variance model to model. Additionally, we find that a large proportion of the most detectable potential shortcuts consist of missing values.}
\label{fig:detect_vs_split}
\end{figure}

As shown in Figure~\ref{fig:detect_vs_split}, we find that, across model classes, G-AUDIT-based detectability correlates strongly  with SPLIT-measured recoverability, with Spearman's \(\rho\) values of .92, .92, .76, .84, .75, and .86 for decision tree, random forest, logistic regression, FT-Transformer, XGBoost, and naive Bayes task models respectively (all \(p < .0001\)). Importantly, both obvious shortcuts such as 'temperature missing'—which is easily detectable because it is explicitly represented in the input data as a negative placeholder value for temperature—as well as less obvious instances like dopamine, norepinephrine, vasopressin, ventilator and IV usage are highly detectable. We do note some variability in performance that does not always seem correlated with model strength as measured in Figure~\ref{fig:sepsis_task_performance}. This suggests using a small ensemble of models of varying complexity may provide a more holistic view of detectability.

\begin{figure}[h]
    \centering
\includegraphics[width=\textwidth]{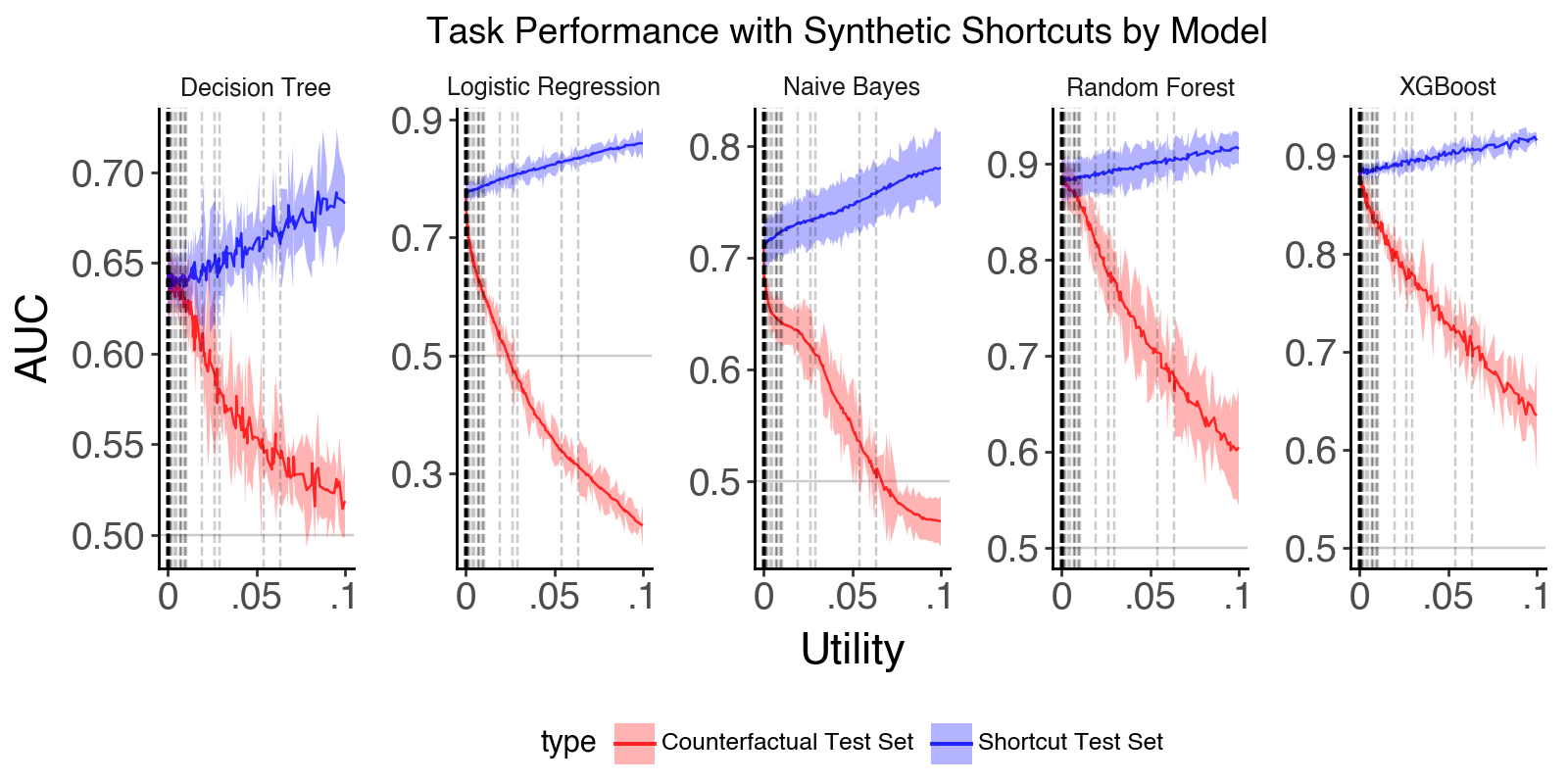}
    \caption{Following our calibration procedure, we can explore the worst-case impact a completely detectable shortcut has on downstream models. Dotted vertical lines correspond to the utility values of real attributes in the dataset. Blue lines correspond to test sets where the shortcut's relationship to the label holds true; red correspond to the worst-case counterfactual set. }
\label{fig:shortcut_vs_counterfactual}
\end{figure}

By running our synthetic calibration process (Sec.~\ref{subsec:bounding_performance}), we identify that some of the attributes in our set could cause worst case performance drops as great as ~0.2 AUC for an average task model assuming full detectability (see Figure~\ref{fig:shortcut_vs_counterfactual}). 

%% file: sections/Discussion.tex
\section{Discussion}
The G-AUDIT provides a procedural means to identify and quantify potential sources of dataset bias and machine learning shortcuts.  The technique identifies potentially harmful relationships between attributes and task labels that may be exploited via the input data.  G-AUDIT estimates the presence and strength of these relationships through attribute utility and detectability.  Attributes with both high-utility and high-detectability are indicative of dataset bias and act as primary targets for downstream algorithmic auditing.

G-AUDIT is the first known method for fully quantitative measurement of dataset bias relative to both clinical and imaging factors. Prior to this method, the closest points of comparison consider only patient attributes (such as age or sex) and overlook other aspects of the dataset which may be a stronger source of bias (such as the collection site or model of imaging device).  In fact, our analysis finds that non-patient attributes are often of greatest concern which is consistent with previous studies that focused on specific shortcut learning scenarios such as the use of digital watermarks in x-ray images~\cite{degrave2021ai} or presence of chest drains for detecting pneumothorax~\cite{oakden2020hidden}.  While these cases were found through manual inspection by researchers, our auditing procedure would automatically and more efficiently enable detection of these possible sources of bias prior to model training. Since our audits are implicitly tied to measurable and interpretable attributes of the data, clinicians and AI developers are better equipped to interpret the results of the audit and identify relevant courses of action (e.g., bias mitigation or model auditing strategies).

We show the generality of G-AUDIT by applying it across multiple machine learning tasks and data modalities.  G-AUDIT is shown to be effective in both causal and anti-causal scenarios where the relationship between the labels and data may be reversed. 
While we can run G-AUDIT easily for both scenarios, additional knowledge or \textit{a priori} assumptions about the directionality can be leveraged to improve the overall efficiency. 
Future work will consider augmenting G-AUDIT to also include automated detection of label-data directionality.  

While G-AUDIT is guaranteed to examine every sample in the dataset, as the the size of the audited datasets or the number of attributes increases, additional considerations will be necessary to ensure that G-AUDIT remains computationally viable.  Nonetheless, unlike algorithmic audits which require the procedure to be run for each new task model, G-AUDIT can be run once per dataset-task combination and adding new tasks only requires re-calculating utility and detectability given the original attribute predictions and new task labels.

In this work, we provide a generalized, quantitative technique for generating hypotheses related to the non-representativeness of a dataset to inform downstream model training and auditing.  As adoption of data-driven machine learning methods for safety- and cost-critical medical applications continues to grow,  principled quantitative methods for analyzing the underlying training and evaluation data are important tools for highlighting potential limitations of the data.   G-AUDIT provides the first such approach and results demonstrate that commonly overlooked dataset attributes may induce biased AI outputs that negatively impact the diagnosis and treatment of underlying patient conditions.  Our method provides a positive step towards identifying and mitigating these risks.

%% file: sections/Methods.tex
\section{Methods}
\label{sec:methods}
\subsection{Utility and Detectability}
The core objective of our data auditing procedure is to identify and quantify the degree to which each attribute of the data represents a potential learning shortcut. 

For our purposes, datasets consist of samples ($X$), associated metadata/attributes ($A$), and task labels ($Y$). We establish the existence (not direction) of a relationship between $A \leftrightarrow Y$ through a measure we call \textit{utility}.  The utility of an attribute refers to our ability to infer the value of the task label $Y$ simply by observing $A$.  In the extreme case, if $A$ is perfectly correlated with $Y$, then machine learning models simply need to detect $A$ in order to correctly solve the task. 

However, a large value for an attribute's utility is not sufficient to consider it a shortcut risk.  For this, we require to know the attribute's \textit{detectability}.  Since machine learning models will typically only take $X$ (but not $A$) as input, \textit{detectability} measures the extent to which the value of $A$ can be inferred from $X$. 

Attributes with high-utility and high-detectability represent the greatest risk for biasing downstream task models.  However, while we use utility and detectability as a proxy for risk, features that are causally relevant to the task and have high-utility and -detectability are useful, rather than representing possible shortcuts. Therefore, domain expertise is still needed to determine whether high-risk attributes are reasonable features or unanticipated dataset flaws.  

\subsection{Measuring Utility and Detectability}
We use information theory and principles of causal inference to measure the utility and detectability of dataset attributes.  \textit{Utility} is measured as the mutual information, $MI(A; Y) = H(Y) - H(Y|A)$. It represents the reduction in uncertainty about the value of $Y$ by observing $A$ while accounting for the entropy of the underlying distributions and adjusting chance based on the number of categories (as in~\cite{vinh2009ami,vinh2010ami}). We rely on the faithfulness assumption which implies that if there is a relationship between $A,Y$, then $MI(A;Y) > 0$.  

For \textit{detectability}, the objective is to determine the ability to infer the value $A$ from the data and we accomplish this by first training a surrogate model $f: A \rightarrow X$ to predict $A$ from $X$. 
We next estimate the utility and detectability between data and labels for both possible directions of dependency between data and labels.  In many cases, clinician and domain expertise may allow us to eliminate certain directions of dependency thus improving the interpretability of the results and reducing overall computation.
Our ultimate goal is to identify whether there is a relationship $A \leftrightarrow X$ and if so measure its strength. We consider the direction of dependence between data and labels in order to control for the potential of information leaking through $Y$ and this process relies on the assumed/known relationship between $Y, X$.

In anti-causal scenarios (i.e., $Y \rightarrow X$), we know it could be the case that $A\rightarrow Y \rightarrow X$ and as a result information could be leaked about the attribute into $X$ not by a shortcut, but by solving the task itself. 
To control for this, we condition on $Y$ during the training process of our surrogate attribute prediction model $f : X \rightarrow \hat{A}$. Specifically, we assume either $Y$ is already discrete or define $Y^D$ as a sufficiently granular discretization of $Y$. Then, we partition $X, Y^D$ into disjoint training subsets $S$ s.t. for given subset $S_i$, all $y^{D_j} \in S_i$ have the same value. For each subset $S_i \in S$, we train separate surrogates $f_i(X) = \hat{A}$. By training separate surrogates in this manner, we ensure that differences in task labels $Y$ that could be used by $f_i$ to predict a given attribute are minimized. Based on test set predictions obtained using cross-validation with all $f_i$, we obtain a full set of predictions ($\hat{A}$) for the entire dataset and we can similarly measure $MI(A;\hat{A})$ to understand how recoverable $A$ is from $X$ while reducing the impact of task relevant information. 

In the causal case (i.e., $X \rightarrow Y$), we are able to directly estimate the relationship without conditioning. In fact, we cannot condition on $Y$ since $Y$ represents a collider (i.e., $Y$ may be dependent on both $X$ and $A$). In that case, measuring the strength of relationship between $X \leftrightarrow A$ given $Y$ could give falsely inflated results since conditioning on a collider $Y$ creates an otherwise non-existent association between its parents ($X, A$). For example, if two conditions both increase mortality rate through different pathways, then given that a patient has died, knowledge of either attribute provides information about the other (e.g., given $A$ is the presence or absence of trauma wounds and $X$ contains information relating to disease, a patient who died but did not have trauma wounds is more likely to have had disease: $A\leftrightarrow X | Y$ even though  $A\perp X$ ). As a result, we do not condition for cases where we expect $X \rightarrow Y$ and instead directly estimate $MI(A;\hat{A})$.

\subsection{Data Audit Procedure}
A key aspect of the data auditing procedure is that every sample in the dataset contributes to the calculation of attribute utility and detectability.  For determining utility, only the labels and metadata are required and can be estimated without any model training.  However, detectability requires attribute prediction, so we ensure that every sample in the dataset contributes to the detectability estimate using a cross-validation procedure that ensures unbiased predictions of $\hat{A}$.  

We first partition the full dataset into $K$ disjoint folds.  For each fold, we hold out the data of that fold for testing and use the data from the $K-1$ remaining folds to train or finetune a sufficiently expressive machine learning model to predict $\hat{A}$ from $X$.  Given that trained model, we predict $\hat{A}$ on the held-out fold's data.  After repeating the train/test procedure for each fold, and, as necessary due to the dependency between attribute and label, each value in $Y^D$, we aggregate all predictions together to compute detectability measures.

\subsection{Bounding Performance Risk}
\label{subsec:bounding_performance}
While utility and detectability measures provide a quantitative measure of bias, how to interpret them in the context of potential model performance risks is less clear. The magnitudes of these measures are not easily comparable across datasets so we cannot rely on thresholds or conventions to directly assess risk. We could instead look to the relative rankings of attributes with respect to utility/detectability to understand attribute risk ordering, but cannot directly translate values to drops in task-relevant metrics like AUC.

To address this concern, we provide a supplementary method for generating an upper bound on performance risk with which attribute utility can be interpreted. To construct the bound, we first create a synthetic attribute which we can insert with 100\% detectability (e.g., a visible watermark in a fixed image position, a single token added to a text input, an added column to tabular data).  We then vary the utility of this synthetic attribute and train a task model for each case.  We evaluate the task models on synthetic and counterfactual data distributions and measure the resulting performance drop between the two distributions.  The counterfactual data distribution is constructed in such a way as to create the worst case scenario where the synthetic attribute is anti-correlated with the true label and any shortcut exploited during training will yield worst-case behavior at test time.

Formally, let $X$ be the feature data, $Y$ be the true binary task labels (of length $N_Y$), and $A$ be the synthetic artifact values.  For simplicity, we assume $A$ is also binary and initialize the values of $A$ to have the same values as $Y$ in the dataset (i.e., the normalized utility would be $1.0$ when $A=Y$).  To vary the utility, we randomly select an index set $\gI$ for $N < \frac{N_Y}{2}$ rows of $A$ and flip the values in those rows (i.e., if $i \in \gI$ then $A[i] = \neg Y[i]$). This preserves the existence of the relationship between $Y,A$ but reduces its strength.  For each $i \in \gI$, we also insert a fully detectable artifact into $X$.  We run the same training and testing procedure as described in Section~\ref{sub:train-test} to get the baseline task performance numbers for the synthetic distribution.

To construct the counterfactual distribution, we create an additional test set as follows. For each image in the test set, we insert the synthetic artifact $A_C$ to be anti-correlated with the label such that $Y=0 \Rightarrow A_C=1$ and vice versa. In the worst case, if the task model exploited the synthetic artifact shortcut during training, then at test time, it will be more likely to predict $\hat{Y}=1$ when $A_C=1$ resulting in more errors and lower AUC.  As the utility of $A$ increases in the training set, the risk of performance degradation in this context also increases.  

We view this as an approximate upper bound on risk because of the fact that we have controlled for the detectability of the synthetic attribute and thus can understand the performance risks associated with each value of utility.  To assess risk for attributes in the original dataset, we can look at the worst-case AUC drop relative to the measured utility for the original attributes.

\subsection{Model Training and Prediction}
\label{sub:train-test}
\subsubsection{Skin Lesion Classification}
For the attribute prediction step of our auditing procedure, we train ResNet18~\cite{he2016deep} networks since they are sufficiently expressive for determining attribute detectability but not as prone to overfitting or training instability as larger or more complex architectures. For task label prediction, we train a more expressive SwinT~\cite{liu2021swin} (using RandAugment~\cite{Cubuk2019-kv}) that is capable of solving the more challenging vision task and demonstrates that the detectability results produced by our auditing procedure generalize to more complex architectures. 

For continuous attributes (e.g., age, image height/width), we discretize the attribute values and train the attribute predictor in a multi-class setting.  For instance, for age, we take $y = \lfloor \frac{age}{5} \rfloor$ which yields 18 total classes for the age attribute predictor.  We use a similar binning procedure for the image height, width, and year.

All prediction networks are pretrained on ImageNet with weights provided by the popular \verb|torchvision| package~\cite{torchvision2016}. For computing G-AUDIT's detectability, networks are fine-tuned for 10 epochs using the AdamW optimizer~\cite{loshchilov2017decoupled} with a learning rate of 5e-5, weight decay of 0.01, momentum parameters $(0.9, 0.999)$, and linear learning rate decay with $\gamma = 0.7$.  Cross-entropy loss is used for training both attribute prediction and task models. All images are resized to $(224, 224)$ and normalized using standard ImageNet mean/std statistics. Horizontal/vertical flips are only applied during attribute predictor network training.  

\subsubsection{Stigmatizing Language in Electronic Health Records}
We fine-tune BERT \cite{devlin2019bert} models for both attribute prediction and each of the three clinical tasks: (1) compliance, (2) appearance and demeanor, and (3) credibility and obstinacy. Attributes we evaluate as potential shortcuts include patient race, sex, and clinical specialty (see Appendix \ref{subsec:ehr_attributes}). The dataset consists of manually annotated samples with a context window of 10 words before and 10 after each identified potentially stigmatizing anchor word~\cite{harrigian2023characterization}. The full list of anchor words for each task is available in Appendix~\ref{subsec:anchor_words}. As each task has a unique set of anchors, the input and attribute metadata are different across tasks.     
Hyperparameters for all task and attribute models are held constant. We use AdamW~\cite{loshchilov2017decoupled} with a fixed learning rate of 5e-5 and weight decay of 1e-5, a batch size of 16, dropout with probability 0.1, 10 training epochs with early stopping, and class balanced cross-entropy loss for all experiments.

\subsubsection{Mortality Prediction from Intensive Care Unit Data}
Our task models are FT-Transformers \cite{gorishniy2021FtTransformer} trained with default hyperparameters to predict mortality. For attribute prediction, we compare logistic regression, decision tree, Random Forest~\cite{breiman2001random}, FT-Transformer, naive bayes, and XGBoost~\cite{chen2016xgboost} models. We discretize continuous valued attributes prior to calculating mutual information-based estimators. We select the number of bins for discretization automatically via the Freedman-Diaconis rule \cite{freedman1981bin}. To ensure sufficient examples across cross-validation splits, we combine and drop categories of attributes where necessary when there are fewer than 100 members having the value within our dataset (e.g. ethnicities \verb|Guatemalan| and \verb|Honduran|).

\subsubsection{SPLIT method}
\label{subsub:split}
As an alternate baseline form of estimating detectability and following~\cite{Gichoya2022ReadingRace}, we test whether task model representations implicitly encode attribute information.  Given a pre-trained classifier, we remove the final fully-connected layer and replace it with a randomly initialized linear layer.  We freeze the weights of the pre-trained network and finetune only the linear layer to predict the specified attribute value.  As before, we perform cross-validation and measure the performance of the finetuned model on the aggregated predictions across all folds.  Better than chance performance is an indicator that model representations encode some degree of attribute information.  However, it does not necessarily indicate that the dataset itself is biased as some attribute information may be relevant to solving the task even when the dataset is balanced with respect to the attribute itself.

%% file: sections/Supplemental.tex
\appendix
\section{EHR Dataset}
\label{subsec:ehr_attributes}

\subsubsection{EHR Tasks:}
The Johns Hopkins Medicine (JHM) dataset we use seeks to enable the prediction of three types of stigma in healthcare. For each of these three tasks, ground truth labels from the initial work were formed in a two-step process. First, a set of anchor words specific to each task were identified in the raw note text. Data samples were created that consist of ten words to the left and right of each identified anchor word. Examples were labeled as to whether they represented a case of stigmatizing language (e.g. 'the patient \textit{claims} they were..' or not, 'the patient's \textit{claims} were denied'). The team that originally annotated each example consisted of one research assistant and several physician coauthors. At least two annotators independently labeled each example. 

The three tasks and associated anchor words are:
\label{subsec:anchor_words}
\begin{enumerate}
\item \textbf{Credibility \& Obstinacy.} Physician doubt regarding patient testimony or descriptions of patients as obstinate. \paragraph{Anchor Words:}
adamant, adamantly, adament, adamently, claim, claimed, claiming, claims, insist, insisted, insistence, insisting, insists \\
\item \textbf{Compliance.} Related to whether or not patients appear to follow medical advice. \paragraph{Anchor Words:}
Adherance, adhere, adhered, adherence, adherent, adheres, adhering, compliance, compliant, complied, complies, comply, complying, declined, declines, declining, nonadherance, nonadherence, nonadherent, noncompliance, noncompliant, refusal, refuse, refused, refuses, refusing \\ 
\item \textbf{Descriptions of Appearance / Demeanor.} A description of the patient's appearance and/or behavior. \paragraph{Anchor Words:} 
Aggression, aggressive, aggression, aggressive, aggressively, agitated, agitation, anger, angered, angers, angrier, angrily, angry, argumentative, argumentatively, belligerence, belligerent, belligerently, charming, combative, combatively, confrontational, cooperative, defensive, delightful, disheveled, drug seeking, drug-seeking, exaggerate, exaggerates, exaggerating, historian, lovely, malinger, malingered, malingerer, malingering, malingers, narcotic seeking, narcotic-seeking, pleasant, pleasantly, poorly groomed, poorly-groomed, secondary gain, uncooperative, unkempt, unmotivated, unwilling, unwillingly, well groomed.
\end{enumerate} 

\section{Potential Shortcut Attributes}
\label{subsec:shortcuts}

We next outline the attributes evaluated as potential shortcuts via our G-AUDIT method.

\subsection{Skin lesion classification}
Dataset audits for the skin lesion classification task assessed the following attributes as candidate shortcuts: age, anatomical location, image height/width, sex, skin color (Fitzpatrick scale), year

\subsection{Stigmatizing language in EHR data}
For each of the EHR tasks, we have three potential shortcut attributes available from the original patient visit metadata. These are patient sex, race, and the visit's clinical specialty within the JHM hospital system. Clinical specialties can be either: Internal Medicine, Surgery, Emergency Medicine, OB-GYN, or Pediatrics. 

\subsection{Mortality Prediction from ICU Data} 
Potential shortcuts for the mortality prediction task are one-hot encoded. Categories with fewer than 100 examples are merged wherever possible.

\paragraph{Ethnicity attributes:}
ethnicity-Asian, ethnicity-Asian - Chinese, ethnicity-Black/African American, ethnicity-Black/Cape Verdean, ethnicity-Hispanic OR Latino, ethnicity-Hispanic/Latino - Puerto Rican, ethnicity-White, ethnicity-Other, ethnicity-Patient declined to answer, ethnicity-Unable to obtain, ethnicity-Unknown/not specified.

\paragraph{Insurance attributes:}
insurance-Government, insurance-Medicaid, insurance-Medicare, insurance-Private, insurance-Self Pay. 

\paragraph{Intervention attributes:}
vent, vaso, dobutamine, dopamine, epinephrine, milrinone, norepinephrine, phenylephrine, vasopressin, colloid-bolus, crystalloid-bolus, nivdurations.

\paragraph{Missing data attributes:} 
heart rate missing, systolic blood pressure missing, temperature missing, blood urea nitrogen missing, white blood cell count missing, potassium missing, sodium missing, bicarbonate missing, bilirubin missing, glascow coma scale total missing, partial pressure of oxygen missing, fraction inspired oxygen missing.

%% file: main.bbl
\begin{thebibliography}{56}
\providecommand{\natexlab}[1]{#1}
\providecommand{\url}[1]{\texttt{#1}}
\expandafter\ifx\csname urlstyle\endcsname\relax
  \providecommand{\doi}[1]{doi: #1}\else
  \providecommand{\doi}{doi: \begingroup \urlstyle{rm}\Url}\fi

\bibitem[Aka et~al.(2021)Aka, Burke, Bauerle, Greer, and Mitchell]{Aka2021-kq}
O.~Aka, K.~Burke, A.~Bauerle, C.~Greer, and M.~Mitchell.
\newblock Measuring model biases in the absence of ground truth.
\newblock In \emph{{AAAI/ACM} AIES}. ACM, 2021.

\bibitem[Bertsimas et~al.(2021)Bertsimas, Mingardi, and Stellato]{bertsimas2021machine}
D.~Bertsimas, L.~Mingardi, and B.~Stellato.
\newblock Machine learning for real-time heart disease prediction.
\newblock \emph{IEEE Journal of Biomedical and Health Informatics}, 25\penalty0 (9):\penalty0 3627--3637, 2021.

\bibitem[Breiman(2001)]{breiman2001random}
L.~Breiman.
\newblock Random forests.
\newblock \emph{Machine learning}, 45:\penalty0 5--32, 2001.

\bibitem[Chambon et~al.(2024)Chambon, Delbrouck, Sounack, Huang, Chen, Varma, Truong, Chuong, and Langlotz]{Chambon2024-vz}
P.~Chambon, J.-B. Delbrouck, T.~Sounack, S.-C. Huang, Z.~Chen, M.~Varma, S.~Q.~H. Truong, C.~T. Chuong, and C.~P. Langlotz.
\newblock {CheXpert} plus: Augmenting a large chest {X}-ray dataset with text radiology reports, patient demographics and additional image formats.
\newblock \emph{arXiv [cs.CL]}, May 2024.

\bibitem[Chen and Guestrin(2016)]{chen2016xgboost}
T.~Chen and C.~Guestrin.
\newblock Xgboost: A scalable tree boosting system.
\newblock In \emph{Proceedings of the 22nd acm sigkdd international conference on knowledge discovery and data mining}, pages 785--794, 2016.

\bibitem[Codella et~al.(2018)Codella, Gutman, Celebi, Helba, Marchetti, Dusza, Kalloo, Liopyris, Mishra, Kittler, et~al.]{codella2018skin}
N.~C. Codella, D.~Gutman, M.~E. Celebi, B.~Helba, M.~A. Marchetti, S.~W. Dusza, A.~Kalloo, K.~Liopyris, N.~Mishra, H.~Kittler, et~al.
\newblock Skin lesion analysis toward melanoma detection: A challenge at the 2017 international symposium on biomedical imaging (isbi), hosted by the international skin imaging collaboration (isic).
\newblock In \emph{2018 IEEE 15th international symposium on biomedical imaging (ISBI 2018)}, pages 168--172. IEEE, 2018.

\bibitem[Combalia et~al.(2019)Combalia, Codella, Rotemberg, Helba, Vilaplana, Reiter, Carrera, Barreiro, Halpern, Puig, et~al.]{combalia2019bcn20000}
M.~Combalia, N.~C. Codella, V.~Rotemberg, B.~Helba, V.~Vilaplana, O.~Reiter, C.~Carrera, A.~Barreiro, A.~C. Halpern, S.~Puig, et~al.
\newblock Bcn20000: Dermoscopic lesions in the wild.
\newblock \emph{arXiv preprint arXiv:1908.02288}, 2019.

\bibitem[Cubuk et~al.(2019)Cubuk, Zoph, Shlens, and Le]{Cubuk2019-kv}
E.~D. Cubuk, B.~Zoph, J.~Shlens, and Q.~V. Le.
\newblock {{RandAugment}}: Practical automated data augmentation with a reduced search space.
\newblock Sept. 2019.

\bibitem[DeGrave et~al.(2021)DeGrave, Janizek, and Lee]{degrave2021ai}
A.~J. DeGrave, J.~D. Janizek, and S.-I. Lee.
\newblock Ai for radiographic covid-19 detection selects shortcuts over signal.
\newblock \emph{Nature Machine Intelligence}, 2021.

\bibitem[Devlin et~al.(2019)Devlin, Chang, Lee, and Toutanova]{devlin2019bert}
J.~Devlin, M.~Chang, K.~Lee, and K.~Toutanova.
\newblock {BERT:} pre-training of deep bidirectional transformers for language understanding.
\newblock In J.~Burstein, C.~Doran, and T.~Solorio, editors, \emph{Proceedings of the 2019 Conference of the North American Chapter of the Association for Computational Linguistics: Human Language Technologies, {NAACL-HLT} 2019, Minneapolis, MN, USA, June 2-7, 2019, Volume 1 (Long and Short Papers)}, pages 4171--4186. Association for Computational Linguistics, 2019.
\newblock \doi{10.18653/V1/N19-1423}.
\newblock URL \url{https://doi.org/10.18653/v1/n19-1423}.

\bibitem[Dreizin et~al.(2022)Dreizin, Nixon, Hu, Albert, Yan, Yang, Chen, Liang, Kim, Jeudy, et~al.]{dreizin2022pilot}
D.~Dreizin, B.~Nixon, J.~Hu, B.~Albert, C.~Yan, G.~Yang, H.~Chen, Y.~Liang, N.~Kim, J.~Jeudy, et~al.
\newblock A pilot study of deep learning-based ct volumetry for traumatic hemothorax.
\newblock \emph{Emergency radiology}, 29\penalty0 (6):\penalty0 995--1002, 2022.

\bibitem[Drenkow et~al.(2021)Drenkow, Sani, Shpitser, and Unberath]{drenkow2021systematic}
N.~Drenkow, N.~Sani, I.~Shpitser, and M.~Unberath.
\newblock A systematic review of robustness in deep learning for computer vision: Mind the gap?
\newblock \emph{arXiv preprint arXiv:2112.00639}, 2021.

\bibitem[Efron and Tibshirani(1993)]{efron1993bootstrap}
B.~Efron and R.~J. Tibshirani.
\newblock An introduction to the bootstrap chapman \& hall.
\newblock \emph{New York}, 436, 1993.

\bibitem[Freedman and Diaconis(1981)]{freedman1981bin}
D.~Freedman and P.~Diaconis.
\newblock On the histogram as a density estimator:{L2} theory.
\newblock \emph{Zeitschrift für Wahrscheinlichkeitstheorie und Verwandte Gebiete}, 57\penalty0 (4):\penalty0 453--476, Dec. 1981.
\newblock ISSN 1432-2064.
\newblock \doi{10.1007/BF01025868}.
\newblock URL \url{https://doi.org/10.1007/BF01025868}.

\bibitem[Geirhos et~al.(2020)Geirhos, Jacobsen, Michaelis, Zemel, Brendel, Bethge, and Wichmann]{geirhos2020shortcut}
R.~Geirhos, J.-H. Jacobsen, C.~Michaelis, R.~Zemel, W.~Brendel, M.~Bethge, and F.~A. Wichmann.
\newblock Shortcut learning in deep neural networks.
\newblock \emph{Nature Machine Intelligence}, 2020.

\bibitem[Gichoya et~al.(2022)Gichoya, Banerjee, Bhimireddy, Burns, Celi, Chen, Correa, Dullerud, Ghassemi, Huang, Kuo, Lungren, Palmer, Price, Purkayastha, Pyrros, Oakden-Rayner, Okechukwu, Seyyed-Kalantari, Trivedi, Wang, Zaiman, and Zhang]{Gichoya2022ReadingRace}
J.~W. Gichoya, I.~Banerjee, A.~R. Bhimireddy, J.~L. Burns, L.~A. Celi, L.-C. Chen, R.~Correa, N.~Dullerud, M.~Ghassemi, S.-C. Huang, P.-C. Kuo, M.~P. Lungren, L.~J. Palmer, B.~J. Price, S.~Purkayastha, A.~T. Pyrros, L.~Oakden-Rayner, C.~Okechukwu, L.~Seyyed-Kalantari, H.~Trivedi, R.~Wang, Z.~Zaiman, and H.~Zhang.
\newblock Ai recognition of patient race in medical imaging: a modelling study.
\newblock \emph{The Lancet Digital Health}, 2022.

\bibitem[Gichoya et~al.(2023)Gichoya, Thomas, Celi, Safdar, Banerjee, Banja, Seyyed-Kalantari, Trivedi, and Purkayastha]{Gichoya2023-gf}
J.~W. Gichoya, K.~Thomas, L.~A. Celi, N.~Safdar, I.~Banerjee, J.~D. Banja, L.~Seyyed-Kalantari, H.~Trivedi, and S.~Purkayastha.
\newblock {AI} pitfalls and what not to do: mitigating bias in {AI}.
\newblock \emph{Br. J. Radiol.}, 96\penalty0 (1150):\penalty0 20230023, Oct. 2023.

\bibitem[Glocker et~al.(2022)Glocker, Jones, Bernhardt, and Winzeck]{Glocker2022-st}
B.~Glocker, C.~Jones, M.~Bernhardt, and S.~Winzeck.
\newblock Risk of bias in chest x-ray foundation models.
\newblock Sept. 2022.

\bibitem[Glocker et~al.(2023)Glocker, Jones, Bernhardt, and Winzeck]{glocker2023algorithmic}
B.~Glocker, C.~Jones, M.~Bernhardt, and S.~Winzeck.
\newblock Algorithmic encoding of protected characteristics in chest {X}-ray disease detection models.
\newblock \emph{eBioMedicine}, 89:\penalty0 104467, Mar. 2023.
\newblock ISSN 23523964.
\newblock \doi{10.1016/j.ebiom.2023.104467}.
\newblock URL \url{https://linkinghub.elsevier.com/retrieve/pii/S2352396423000324}.

\bibitem[Gorishniy et~al.(2021)Gorishniy, Rubachev, Khrulkov, and Babenko]{gorishniy2021FtTransformer}
Y.~Gorishniy, I.~Rubachev, V.~Khrulkov, and A.~Babenko.
\newblock Revisiting deep learning models for tabular data.
\newblock In M.~Ranzato, A.~Beygelzimer, Y.~Dauphin, P.~Liang, and J.~W. Vaughan, editors, \emph{Advances in Neural Information Processing Systems}, volume~34, pages 18932--18943. Curran Associates, Inc., 2021.
\newblock URL \url{https://proceedings.neurips.cc/paper_files/paper/2021/file/9d86d83f925f2149e9edb0ac3b49229c-Paper.pdf}.

\bibitem[Harrigian et~al.(2023)Harrigian, Zirikly, Chee, Ahmad, Links, Saha, Beach, and Dredze]{harrigian2023characterization}
K.~Harrigian, A.~Zirikly, B.~Chee, A.~Ahmad, A.~Links, S.~Saha, M.~C. Beach, and M.~Dredze.
\newblock Characterization of stigmatizing language in medical records.
\newblock In \emph{Proceedings of the 61st Annual Meeting of the Association for Computational Linguistics (Volume 2: Short Papers)}, pages 312--329, 2023.

\bibitem[He et~al.(2016)He, Zhang, Ren, and Sun]{he2016deep}
K.~He, X.~Zhang, S.~Ren, and J.~Sun.
\newblock Deep residual learning for image recognition.
\newblock In \emph{IEEE/CVPR}, pages 770--778, 2016.

\bibitem[Henry~Hinnefeld et~al.(2018)Henry~Hinnefeld, Cooman, Mammo, and Deese]{Henry_Hinnefeld2018-vi}
J.~Henry~Hinnefeld, P.~Cooman, N.~Mammo, and R.~Deese.
\newblock Evaluating fairness metrics in the presence of dataset bias.
\newblock Sept. 2018.

\bibitem[Herbert et~al.(2023)Herbert, Hou, Bradley, Hager, Boland, Ramulu, Unberath, and Yohannan]{herbert2023forecasting}
P.~Herbert, K.~Hou, C.~Bradley, G.~Hager, M.~V. Boland, P.~Ramulu, M.~Unberath, and J.~Yohannan.
\newblock Forecasting risk of future rapid glaucoma worsening using early visual field, oct, and clinical data.
\newblock \emph{Ophthalmology Glaucoma}, 6\penalty0 (5):\penalty0 466--473, 2023.

\bibitem[Jabbour et~al.(2020)Jabbour, Fouhey, Kazerooni, Sjoding, and Wiens]{jabbour2020deep}
S.~Jabbour, D.~Fouhey, E.~Kazerooni, M.~W. Sjoding, and J.~Wiens.
\newblock Deep learning applied to chest x-rays: Exploiting and preventing shortcuts.
\newblock In \emph{Machine Learning for Healthcare Conference}, pages 750--782. PMLR, 2020.

\bibitem[Jaspers et~al.(2024)Jaspers, Boers, Kusters, Jong, Jukema, de~Groof, Bergman, de~With, and van~der Sommen]{Jaspers2024-vv}
T.~J.~M. Jaspers, T.~G.~W. Boers, C.~H.~J. Kusters, M.~R. Jong, J.~B. Jukema, A.~J. de~Groof, J.~J. Bergman, P.~H.~N. de~With, and F.~van~der Sommen.
\newblock Robustness evaluation of deep neural networks for endoscopic image analysis: Insights and strategies.
\newblock \emph{Med. Image Anal.}, 94:\penalty0 103157, May 2024.

\bibitem[Johnson et~al.(2016)Johnson, Pollard, Shen, Lehman, Feng, Ghassemi, Moody, Szolovits, Anthony~Celi, and Mark]{johnson2016mimic}
A.~E. Johnson, T.~J. Pollard, L.~Shen, L.-w.~H. Lehman, M.~Feng, M.~Ghassemi, B.~Moody, P.~Szolovits, L.~Anthony~Celi, and R.~G. Mark.
\newblock Mimic-iii, a freely accessible critical care database.
\newblock \emph{Scientific data}, 3\penalty0 (1):\penalty0 1--9, 2016.

\bibitem[Johnson et~al.(2023)Johnson, Bulgarelli, Shen, Gayles, Shammout, Horng, Pollard, Hao, Moody, Gow, Lehman, Celi, and Mark]{Johnson2023-mf}
A.~E.~W. Johnson, L.~Bulgarelli, L.~Shen, A.~Gayles, A.~Shammout, S.~Horng, T.~J. Pollard, S.~Hao, B.~Moody, B.~Gow, L.-W.~H. Lehman, L.~A. Celi, and R.~G. Mark.
\newblock {MIMIC}-{IV}, a freely accessible electronic health record dataset.
\newblock \emph{Sci. Data}, 10\penalty0 (1):\penalty0 1, Jan. 2023.

\bibitem[Jones et~al.(2024)Jones, Castro, De~Sousa~Ribeiro, Oktay, McCradden, and Glocker]{Jones2024-na}
C.~Jones, D.~C. Castro, F.~De~Sousa~Ribeiro, O.~Oktay, M.~McCradden, and B.~Glocker.
\newblock A causal perspective on dataset bias in machine learning for medical imaging.
\newblock \emph{Nature Machine Intelligence}, 6\penalty0 (2):\penalty0 138--146, Feb. 2024.

\bibitem[Le~Gall et~al.(1993)Le~Gall, Lemeshow, and Saulnier]{gallPhysiology1993}
J.-R. Le~Gall, S.~Lemeshow, and F.~Saulnier.
\newblock {A New Simplified Acute Physiology Score (SAPS II) Based on a European/North American Multicenter Study}.
\newblock \emph{JAMA}, 270\penalty0 (24):\penalty0 2957--2963, 12 1993.
\newblock ISSN 0098-7484.
\newblock \doi{10.1001/jama.1993.03510240069035}.
\newblock URL \url{https://doi.org/10.1001/jama.1993.03510240069035}.

\bibitem[Liu et~al.(2019)Liu, Faes, Kale, Wagner, Fu, Bruynseels, Mahendiran, Moraes, Shamdas, Kern, et~al.]{liu2019comparison}
X.~Liu, L.~Faes, A.~U. Kale, S.~K. Wagner, D.~J. Fu, A.~Bruynseels, T.~Mahendiran, G.~Moraes, M.~Shamdas, C.~Kern, et~al.
\newblock A comparison of deep learning performance against health-care professionals in detecting diseases from medical imaging: a systematic review and meta-analysis.
\newblock \emph{The lancet digital health}, 1\penalty0 (6):\penalty0 e271--e297, 2019.

\bibitem[Liu et~al.(2021)Liu, Lin, Cao, Hu, Wei, Zhang, Lin, and Guo]{liu2021swin}
Z.~Liu, Y.~Lin, Y.~Cao, H.~Hu, Y.~Wei, Z.~Zhang, S.~Lin, and B.~Guo.
\newblock Swin transformer: Hierarchical vision transformer using shifted windows.
\newblock In \emph{IEEE/CVPR}, 2021.

\bibitem[Lopez et~al.(2017)Lopez, Giro-i Nieto, Burdick, and Marques]{lopez2017skin}
A.~R. Lopez, X.~Giro-i Nieto, J.~Burdick, and O.~Marques.
\newblock Skin lesion classification from dermoscopic images using deep learning techniques.
\newblock In \emph{2017 13th IASTED international conference on biomedical engineering (BioMed)}, pages 49--54. IEEE, 2017.

\bibitem[Loshchilov and Hutter(2017)]{loshchilov2017decoupled}
I.~Loshchilov and F.~Hutter.
\newblock Decoupled weight decay regularization.
\newblock \emph{arXiv preprint arXiv:1711.05101}, 2017.

\bibitem[Mahbod et~al.(2019)Mahbod, Schaefer, Wang, Ecker, and Ellinge]{mahbod2019skin}
A.~Mahbod, G.~Schaefer, C.~Wang, R.~Ecker, and I.~Ellinge.
\newblock Skin lesion classification using hybrid deep neural networks.
\newblock In \emph{ICASSP 2019-2019 IEEE international conference on acoustics, speech and signal processing (ICASSP)}, pages 1229--1233. IEEE, 2019.

\bibitem[maintainers and contributors(2016)]{torchvision2016}
T.~maintainers and contributors.
\newblock Torchvision: Pytorch's computer vision library.
\newblock \url{https://github.com/pytorch/vision}, 2016.

\bibitem[Matsoukas et~al.(2023)Matsoukas, Morey, Lock, Chada, Shigematsu, Marayati, Delman, Doshi, Majidi, De~Leacy, et~al.]{matsoukas2023ai}
S.~Matsoukas, J.~Morey, G.~Lock, D.~Chada, T.~Shigematsu, N.~F. Marayati, B.~N. Delman, A.~Doshi, S.~Majidi, R.~De~Leacy, et~al.
\newblock Ai software detection of large vessel occlusion stroke on ct angiography: a real-world prospective diagnostic test accuracy study.
\newblock \emph{Journal of Neurointerventional Surgery}, 15\penalty0 (1):\penalty0 52--56, 2023.

\bibitem[Nauta et~al.(2021)Nauta, Walsh, Dubowski, and Seifert]{nauta2021uncovering}
M.~Nauta, R.~Walsh, A.~Dubowski, and C.~Seifert.
\newblock Uncovering and correcting shortcut learning in machine learning models for skin cancer diagnosis.
\newblock \emph{Diagnostics}, 2021.

\bibitem[Oakden-Rayner et~al.(2020)Oakden-Rayner, Dunnmon, Carneiro, and R{\'e}]{oakden2020hidden}
L.~Oakden-Rayner, J.~Dunnmon, G.~Carneiro, and C.~R{\'e}.
\newblock Hidden stratification causes clinically meaningful failures in machine learning for medical imaging.
\newblock In \emph{Proc. of the ACM conference on health, inference, and learning}, pages 151--159, 2020.

\bibitem[O'Brien et~al.(2022)O'Brien, Bukowski, Hager, Pezeshk, and Unberath]{o2022evaluating}
M.~O'Brien, J.~Bukowski, G.~Hager, A.~Pezeshk, and M.~Unberath.
\newblock Evaluating neural network robustness for melanoma classification using mutual information.
\newblock In \emph{Medical Imaging 2022: Image Processing}. SPIE, 2022.

\bibitem[Ong~Ly et~al.(2024)Ong~Ly, Unnikrishnan, Tadic, Patel, Duhamel, Kandel, Moayedi, Brudno, Hope, Ross, and McIntosh]{Ong_Ly2024-jh}
C.~Ong~Ly, B.~Unnikrishnan, T.~Tadic, T.~Patel, J.~Duhamel, S.~Kandel, Y.~Moayedi, M.~Brudno, A.~Hope, H.~Ross, and C.~McIntosh.
\newblock Shortcut learning in medical {AI} hinders generalization: method for estimating {AI} model generalization without external data.
\newblock \emph{npj Digital Medicine}, 7\penalty0 (1):\penalty0 1--10, May 2024.

\bibitem[Pavlak et~al.(2023)Pavlak, Drenkow, Petrick, Farhangi, and Unberath]{Pavlak2023-km}
M.~F. Pavlak, N.~G. Drenkow, N.~Petrick, M.~M. Farhangi, and M.~Unberath.
\newblock Data {AUDIT}: Identifying attribute utility- and detectability-induced bias in task models.
\newblock \emph{Med. Image Comput. Comput. Assist. Interv.}, pages 442--452, Apr. 2023.

\bibitem[Pickhardt et~al.(2021)Pickhardt, Graffy, Perez, Lubner, Elton, and Summers]{pickhardt2021opportunistic}
P.~J. Pickhardt, P.~M. Graffy, A.~A. Perez, M.~G. Lubner, D.~C. Elton, and R.~M. Summers.
\newblock Opportunistic screening at abdominal ct: use of automated body composition biomarkers for added cardiometabolic value.
\newblock \emph{RadioGraphics}, 41\penalty0 (2):\penalty0 524--542, 2021.

\bibitem[Pirracchio et~al.(2015)Pirracchio, Petersen, Carone, Rigon, Chevret, and van~der Laan]{pirracchio2015mortality}
R.~Pirracchio, M.~L. Petersen, M.~Carone, M.~R. Rigon, S.~Chevret, and M.~J. van~der Laan.
\newblock Mortality prediction in intensive care units with the super icu learner algorithm (sicula): a population-based study.
\newblock \emph{The Lancet Respiratory Medicine}, 3\penalty0 (1):\penalty0 42--52, 2015.

\bibitem[Seyyed-Kalantari et~al.(2021{\natexlab{a}})Seyyed-Kalantari, Liu, McDermott, Chen, and Ghassemi]{Seyyed-Kalantari2021-fa}
L.~Seyyed-Kalantari, G.~Liu, M.~McDermott, I.~Y. Chen, and M.~Ghassemi.
\newblock {CheXclusion}: Fairness gaps in deep chest x-ray classifiers.
\newblock \emph{Pac. Symp. Biocomput.}, 2021{\natexlab{a}}.

\bibitem[Seyyed-Kalantari et~al.(2021{\natexlab{b}})Seyyed-Kalantari, Zhang, McDermott, Chen, and Ghassemi]{Seyyed-Kalantari2021-aj}
L.~Seyyed-Kalantari, H.~Zhang, M.~B.~A. McDermott, I.~Y. Chen, and M.~Ghassemi.
\newblock Underdiagnosis bias of artificial intelligence algorithms applied to chest radiographs in under-served patient populations.
\newblock \emph{Nat. Med.}, 2021{\natexlab{b}}.

\bibitem[Subbaswamy et~al.(2021)Subbaswamy, Adams, and Saria]{Subbaswamy2021-su}
A.~Subbaswamy, R.~Adams, and S.~Saria.
\newblock Evaluating model robustness and stability to dataset shift.
\newblock In A.~Banerjee and K.~Fukumizu, editors, \emph{Proceedings of The 24th International Conference on Artificial Intelligence and Statistics}, volume 130 of \emph{Proceedings of Machine Learning Research}, pages 2611--2619. PMLR, 2021.

\bibitem[Subbaswamy et~al.(2024)Subbaswamy, Sahiner, Petrick, Pai, Adams, Diamond, and Saria]{Subbaswamy2024-cr}
A.~Subbaswamy, B.~Sahiner, N.~Petrick, V.~Pai, R.~Adams, M.~C. Diamond, and S.~Saria.
\newblock A data-driven framework for identifying patient subgroups on which an {AI}/machine learning model may underperform.
\newblock \emph{NPJ Digit. Med.}, 7\penalty0 (1):\penalty0 334, Nov. 2024.

\bibitem[Tschandl et~al.(2018)Tschandl, Rosendahl, and Kittler]{tschandl2018ham10000}
P.~Tschandl, C.~Rosendahl, and H.~Kittler.
\newblock The ham10000 dataset, a large collection of multi-source dermatoscopic images of common pigmented skin lesions.
\newblock \emph{Scientific data}, 2018.

\bibitem[Vinh et~al.(2009)Vinh, Epps, and Bailey]{vinh2009ami}
N.~X. Vinh, J.~Epps, and J.~Bailey.
\newblock Information theoretic measures for clusterings comparison: is a correction for chance necessary?
\newblock In \emph{Proceedings of the 26th Annual International Conference on Machine Learning}, ICML '09, page 1073–1080, New York, NY, USA, 2009. Association for Computing Machinery.
\newblock ISBN 9781605585161.
\newblock \doi{10.1145/1553374.1553511}.
\newblock URL \url{https://doi.org/10.1145/1553374.1553511}.

\bibitem[Vinh et~al.(2010)Vinh, Epps, and Bailey]{vinh2010ami}
N.~X. Vinh, J.~Epps, and J.~Bailey.
\newblock Information theoretic measures for clusterings comparison: Variants, properties, normalization and correction for chance.
\newblock \emph{JMLR}, 2010.

\bibitem[Winkler et~al.(2019)Winkler, Fink, Toberer, Enk, Deinlein, Hofmann-Wellenhof, Thomas, Lallas, Blum, Stolz, et~al.]{winkler2019association}
J.~K. Winkler, C.~Fink, F.~Toberer, A.~Enk, T.~Deinlein, R.~Hofmann-Wellenhof, L.~Thomas, A.~Lallas, A.~Blum, W.~Stolz, et~al.
\newblock Association between surgical skin markings in dermoscopic images and diagnostic performance of a deep learning convolutional neural network for melanoma recognition.
\newblock \emph{JAMA dermatology}, 2019.

\bibitem[Winkler et~al.(2021)Winkler, Sies, Fink, Toberer, Enk, Abassi, Fuchs, and Haenssle]{winkler2021association}
J.~K. Winkler, K.~Sies, C.~Fink, F.~Toberer, A.~Enk, M.~S. Abassi, T.~Fuchs, and H.~A. Haenssle.
\newblock Association between different scale bars in dermoscopic images and diagnostic performance of a market-approved deep learning convolutional neural network for melanoma recognition.
\newblock \emph{European Journal of Cancer}, 2021.

\bibitem[Wong et~al.(2021)Wong, Otles, Donnelly, Krumm, McCullough, DeTroyer-Cooley, Pestrue, Phillips, Konye, Penoza, et~al.]{wong2021external}
A.~Wong, E.~Otles, J.~P. Donnelly, A.~Krumm, J.~McCullough, O.~DeTroyer-Cooley, J.~Pestrue, M.~Phillips, J.~Konye, C.~Penoza, et~al.
\newblock External validation of a widely implemented proprietary sepsis prediction model in hospitalized patients.
\newblock \emph{JAMA internal medicine}, 181\penalty0 (8):\penalty0 1065--1070, 2021.

\bibitem[Zech et~al.(2018)Zech, Badgeley, Liu, Costa, Titano, and Oermann]{zech2018variable}
J.~R. Zech, M.~A. Badgeley, M.~Liu, A.~B. Costa, J.~J. Titano, and E.~K. Oermann.
\newblock Variable generalization performance of a deep learning model to detect pneumonia in chest radiographs: a cross-sectional study.
\newblock \emph{PLoS medicine}, 2018.

\bibitem[Zhang et~al.(2019)Zhang, Xie, Xia, and Shen]{zhang2019attention}
J.~Zhang, Y.~Xie, Y.~Xia, and C.~Shen.
\newblock Attention residual learning for skin lesion classification.
\newblock \emph{IEEE transactions on medical imaging}, 38\penalty0 (9):\penalty0 2092--2103, 2019.

\end{thebibliography}
